\documentclass[12pt]{article}
\usepackage{abstract,amsmath,amssymb,latexsym}
\usepackage{enumitem,epsf}
\usepackage{fullpage,tikz,float}
\usepackage[numbers]{natbib}
\usepackage[pdftex,colorlinks]{hyperref}
\usepackage{algorithm}
\usepackage{algorithmic}

\usepackage{macros}

\newif\ifdraft
\drafttrue

\title{SGD Learns the Conjugate Kernel Class of the Network}

\author{
	\vspace{1cm}
  Amit Daniely\thanks{Google Brain} 
}

\begin{document}
\maketitle
\setcounter{page}{0}

\thispagestyle{empty}
\maketitle

\begin{abstract}
We show that the standard stochastic gradient decent (SGD) algorithm is guaranteed to learn, in polynomial time, a function that is competitive with the best function in the conjugate kernel space of the network, as defined in \citet{daniely2016toward}. The result holds for log-depth networks from a rich family of architectures.
To the best of our knowledge, it is the first polynomial-time guarantee for the standard neural network learning algorithm for networks of depth more that two.

As corollaries, it follows that for neural networks of any depth between $2$ and $\log(n)$, SGD is guaranteed to learn, in polynomial time, constant degree polynomials with polynomially bounded coefficients. Likewise, it follows  that SGD on large enough networks can learn any continuous function (not in polynomial time), complementing classical expressivity results.
\end{abstract}

\newpage

\section{Introduction}

While stochastic gradient decent (SGD) from a random initialization is probably the most popular supervised learning algorithm today, we have very few results that depicts conditions that guarantee its success. Indeed, to the best of our knowledge, \citet{andoni2014learning} provides the only known result of this form, and it is valid in a rather restricted setting. Namely, for depth-2 networks, where the underlying distribution is Gaussian, the algorithm is full gradient decent (rather than SGD), and the task is regression when the learnt function is a constant degree polynomial.

We build on the framework of \citet{daniely2016toward} to establish guarantees on SGD in a rather general setting. 
\citet{daniely2016toward} defined a framework that associates a reproducing kernel to a network architecture. They also connected the kernel to the network via the random initialization. Namely, they showed that right after the random initialization, any function in the kernel space can be approximated by changing the weights of the last layer. The quality of the approximation depends on the size of the network and the norm of the function in the kernel space.

As optimizing the last layer is a convex procedure, the result of \citet{daniely2016toward} intuitively shows that the optimization process starts from a favourable point for learning a function in the conjugate kernel space.
In this paper we verify this intuition.
Namely, for a fairly general family of architectures (that contains fully connected networks and convolutional networks) and supervised learning tasks, we show that if the network is large enough, the learning rate is small enough, and the number of SGD steps is large enough as well, SGD is guaranteed to learn any function in the corresponding kernel space. 
We emphasize that the number of steps and the size of the network are only required to be polynomial (which is best possible) in the relevant parameters -- the norm of the function, the required accuracy parameter ($\epsilon$), and the dimension of the input and the output of the network. Likewise, the result holds for any input distribution.

To evaluate our result, one should understand which functions it guarantee that SGD will learn.
Namely, what functions reside in the conjugate kernel space, how rich it is, and how good those functions are as predictors.
From an empirical perspective, in \citep{daniely2017random}, it is shown that for standard  convolutional networks the conjugate class contains functions whose performance is close to the performance of the function that is actually learned by the network. This is based on experiments on the standard CIFAR-10 dataset. 
From a theoretical perspective, we list below a few implications that demonstrate the richness of the conjugate kernel space. These implications are valid for fully connected networks of any depth between $2$ and $\log(n)$, where $n$ is the input dimension. Likewise, they are also valid for convolutional networks of any depth between $2$ and $\log(n)$, and with constantly many convolutional layers.
\begin{itemize}
\item SGD is guaranteed to learn in polynomial time constant degree polynomials with polynomially bounded coefficients. As a corollary, SGD is guaranteed to learn in polynomial time conjunctions, DNF and CNF formulas with constantly many terms, and DNF and CNF formulas with constantly many literals in each term. These function classes comprise a considerable fraction of the function classes that are known to be poly-time (PAC) learnable by {\em any method}. Exceptions include constant degree polynomial thresholds with no restriction on the coefficients, decision lists and parities.
\item SGD is guaranteed to learn, not necessarily in polynomial time, {\em any} continuous function. This complements classical universal approximation results that show that neural networks can {\em (approximately) express} any continuous function (see \cite{scarselli1998universal} for a survey). Our results strengthen those results and show that networks are not only able to express those functions, but actually guaranteed to {\em learn} them.
\end{itemize}

\subsection{Related work}
\paragraph{Guarantees on SGD.} As noted above, there are very few results that provide polynomial time guarantees for SGD on NN. One notable exception is the work of \citet{andoni2014learning}, that proves a result that is similar to ours, but in a substantially more restricted setting. Concretely, their result holds for depth-2 fully connected networks, as opposed to rather general architecture and constant or logarithmic depth in our case. Likewise, the marginal distribution on the instance space is assumed to be Gaussian or uniform, as opposed to arbitrary in our case. In addition, the algorithm they consider is full gradient decent, which corresponds to SGD with infinitely large mini-batch, as opposed to SGD with arbitrary mini-batch size in our case. Finally, the underlying task is regression in which the target function is a constant degree polynomial, whereas we consider rather general supervised learning setting.

\paragraph{Other polynomial time guarantees on learning deep architectures.} Various recent papers show that poly-time learning is possible in the case that the the learnt function can be realized by a neural network with certain (usually fairly strong) restrictions on the weights~\citep{livni2014computational, zhang2016l1, zhang2015learning, zhang2016convexified}, or under the assumption that the data is generated by a generative model that is derived from the network architecture~\citep{arora2014provable, arora2016provable}. We emphasize that the main difference of those results from our results and the results of \citet{andoni2014learning} is that they do not provide guarantees on the standard SGD learning algorithm. Rather, they show that under those aforementioned conditions, there are {\em some algorithms}, usually very different from SGD on the network, that are able to learn in polynomial time.

\paragraph{Connection to kernels.} As mentioned earlier, our paper builds on \citet{daniely2016toward}, who developed the association of kernels to NN which we rely on. Several previous papers~\citep{mairal2014convolutional, cho2009kernel, rahimi2009weighted, RahimiRe07, neal2012bayesian, williams1997infinite, kar2012random, pennington2015spherical, bach2015equivalence, bach2014breaking,hazan2015steps, anselmi2015deep} investigated such associations, but in a more restricted settings (i.e., for less architectures). Some of those papers~\citep{rahimi2009weighted, RahimiRe07, daniely2016toward, kar2012random, bach2015equivalence, bach2014breaking} also provide measure of concentration results, that show that w.h.p.\ the random initialization of the network's weights is reach enough to approximate the functions in the corresponding kernel space. As a result, these papers provide polynomial time guarantees on the variant of SGD, where only the last layer is trained. We remark that with the exception of~\cite{daniely2016toward}, those results apply just to depth-2 networks.

\subsection{Discussion and future directions}
We next want to place this work in the appropriate learning theoretic context, and to elaborate further on this paper's approach for investigating neural networks. For the sake of concreteness, let us restrict the discussion to binary classification over the Boolean cube. Namely, given examples from a distribution $\cd$ on $\{\pm 1\}^n\times\{0,1\}$, the goal is to learn a function $h:\{\pm 1\}^{n}\to\{0,1\}$ whose 0-1 error, $\cl^{0-1}_\cd(h) = \Pr_{(\x,y)\sim\cd}\left(h(\x)\ne y\right)$, is as small as possible. We will use a bit of terminology. A {\em model} is a distribution $\cd$ on $\{\pm 1\}^{n}\times\{0,1\}$ and a {\em model class} is a 
collection $\cm$ of models. We note that any function class $\ch\subset \{0,1\}^{\{\pm 1\}^n}$ defines a model class, $\cm(\ch)$, consisting of all models $\cd$ such that $\cl^{0-1}_\cd(h)=0$ for some $h\in\ch$.
We define the {\em capacity} of a model class as the minimal number $m$ for which there is an algorithm such that for every $\cd\in\cm$ the following holds. Given $m$ samples from $\cd$, the algorithm is guaranteed to return, w.p.\ $\ge \frac{9}{10}$ over the samples and its internal randomness, a function $h:\{\pm 1\}^n\to\{0,1\}$ with 0-1 error $\le \frac{1}{10}$. We note that for function classes the capacity is the VC dimension, up to a constant factor.

Learning theory analyses learning algorithms via model classes. Concretely, one fixes some model class $\cm$ and show that the algorithm is guaranteed to succeed whenever the underlying model is from $\cm$. 
Often, the connection between the algorithm and the class at hand is very clear. For example, in the case that the model is derived from a function class $\ch$, the algorithm might simply be one that finds a function in $\ch$ that makes no mistake on the given sample. 
The natural choice for a model class for analyzing SGD on NN would be the class of all functions that can be realized by the network, possibly with some reasonable restrictions on the weights. Unfortunately, this approach it is probably doomed to fail, as implied by various computational hardness results~\citep{BlumRi89, KearnsVa94, blum1994weakly, Kharitonov93, KlivansSh06, klivans2007unconditional, daniely2013average, danielySh2014}.

So, what model classes should we consider? With a few isolated exceptions (e.g.\ \cite{bshouty1998learning}) all known efficiently learnable model classes are either a linear model class, or contained in an efficiently learnable linear model class.
Namely, functions classes composed of compositions of some predefined embedding with linear threshold functions, or linear functions over some finite field.

Coming up we new tractable models would be a fascinating progress. Still, as linear function classes are the main tool that learning theory currently has for providing guarantees on learning, it seems natural to try to analyze SGD via linear model classes. Our work follows this line of thought, and we believe that there is much more to achieve via this approach. Concretely, while our bounds are polynomial, the degree of the polynomials is rather large, and possibly much better quantitative bounds can be achieved. To be more concrete, suppose that we consider simple fully connected architecture, with 2-layers, ReLU activation, and $n$ hidden neurons. In this case, the capacity of the model class that our results guarantee that SGD will learn is $\Theta\left(n^{\frac{1}{3}}\right)$. For comparison, the capacity of the class of all functions that are realized by this network is $\Theta\left(n^2\right)$. As a challenge, we encourage the reader to prove that with this architecture (possibly with an activation that is different from the ReLU), SGD is guaranteed to learn {\em some} model class of capacity that is super-linear in $n$.

\section{Preliminaries}
\paragraph{Notation.} We denote
vectors by bold-face letters (e.g.\ $\x$), 
matrices by upper case letters (e.g.\ $W$), and collection of matrices by bold-face upper case letters (e.g.\ $\W$). The $p$-norm of $\x
\in \reals^d$ is denoted by $\|\x\|_p = \left(\sum_{i=1}^d|x_i|^p\right)^{\frac{1}{p}}$. We will also use the convention that $\|\x\|=\|\x\|_2$. For functions $\sigma:\reals\to\reals$
we let
$$
\|\sigma\| \textstyle
	:=\sqrt{\E_{X\sim\cn(0,1)}\sigma^2(X)}
	\; = \sqrt{\frac{1}{\sqrt{2\pi}}
		\int_{-\infty}^\infty \sigma^2(x)e^{-\frac{x^2}{2}}dx} \,.
$$
Let $G=(V,E)$ be a directed acyclic graph. The set of neighbors incoming to
a vertex $v$ is denoted $\IN(v):=\{u\in V\mid uv\in E\}$. We also denote $\deg(v) = |\In(v)|$.
Given weight function $\delta:V\to [0,\infty)$ and $U\subset V$ we let $\delta(U) = \sum_{u\in U}\delta(u)$.
The $d-1$ dimensional sphere is denoted $\sphere^{d-1} =
\{\x\in\reals^d \mid \|\x\|=1\}$. 
We use $[x]_+$ to denote $\max(x,0)$. 

\paragraph{Input space.} Throughout the paper we assume that each example is
a sequence of $n$ elements, each of which is represented as a unit vector. 
Namely, we fix $n$ and take the input space to be
	$\cx=\cx_{n,d}=\left(\sphere^{d-1}\right)^n$.
Each input example is denoted,
\begin{align} \label{eq:coordinates}
	\x=(\x^1,\ldots,\x^n), ~\textwhere \x^i\in \sphere^{d-1} \,.
\end{align}
While this notation is slightly non-standard, it unifies 
input types seen in various domains (see \cite{daniely2016toward}). 

\paragraph{Supervised learning.} The goal in supervised learning is to
devise a mapping from the input space $\cx$ to an output space $\cy$ based on a sample
$S=\{(\x_1,y_1),\ldots,(\x_m,y_m)\}$, where $(\x_i,y_i)\in\cx\times\cy$ 
drawn i.i.d.\ from a distribution $\cd$ over $\cx\times\cy$.
A supervised learning problem is further specified by an output length $k$ and a loss function
$\ell : \reals^k \times \cy \to [0,\infty)$, and the goal is to find a
predictor $h:\cx\to\reals^k$ whose loss,
$\cl_{\cd}(h) := \E_{(\x,y)\sim\cd} \ell(h(\x),y)$, is small.
The {\em empirical} loss $\cl_{S}(h):= \frac 1 m \sum_{i=1}^m
\ell(h(\x_i),y_i)$ is commonly used as a proxy for the loss
$\cl_{\cd}$. When $h$ is defined by a vector $\w$ of parameters, we will use the notations $\cl_\cd(\w) = \cl_\cd(h)$, $\cl_S(\w)=\cl_S(h)$ and $\ell_{(\x,y)}(\w)=\ell(h(\x),y)$.

Regression problems correspond to $k=1$, $\cy=\reals$ and, for
instance, the squared loss $\ell^{\mathrm{square}}(\hat y,y)=(\hat y -y)^2$.
Binary classification is captured by $k=1$, $\cy=\{\pm 1\}$ and, say, the
zero-one loss $\ell^{0-1}(\hat y,y)= \ind[\hat y y \leq 0]$ or the hinge
loss $\ell^\hinge(\hat y,y)=[1-\hat y y]_+$.
Multiclass classification is captured by $k$ being the number of classes, $\cy = [k]$, and, say, the
zero-one loss $\ell^{0-1}(\hat y,y)= \ind[\hat y_y \leq \argmax_{y'} \hat y_{y'}]$ or the logistic
loss $\ell^{\log}(\hat \y,y) = -\log\left(p_y(\hat \y)\right)$ where $\p:\reals^k\to \Delta^{k-1}$ is given by $p_i(\hat \y) = \frac{e^{\hat y_i}}{\sum_{j=1}^ke^{\hat y_j}}$.
A loss $\ell$ is $L$-Lipschitz if for all $y\in\cy$, the function $\ell_y(\hat y) :=\ell(\hat y,y)$ is $L$-Lipschitz. Likewise, it is
convex if $\ell_y$ is convex for every $y\in\cy$.


\paragraph{Neural network learning.} 
We define a {\em neural network} $\cn$
to be a vertices weighted directed acyclic graph (DAG) whose nodes are denoted $V(\cn)$ and edges
$E(\cn)$. The weight function will be denoted by $\delta:V(\cn)\to [0,\infty)$, and its sole role would be to dictate the distribution of the initial weights. We will refer $\cn$'s nodes by {\em neurons}.
Each of non-input neuron, i.e.\ neuron with incoming edges, is associated with an {\em activation} function
$\sigma_v:\reals\to\reals$. In this paper, an activation can
be any function $\sigma:\reals\to\reals$ that is right and left differentiable, square integrable with respect to the Gaussian
measure on $\reals$, and is {\em normalized} in the sense that
$\|\sigma\|=1$. The set of neurons having only incoming edges are called the
output neurons.
To match the setup of supervised learning defined above, a network $\cn$ has
$nd$ input neurons and $k$ output neurons, denoted $o_1,\ldots,o_k$. A
network $\cn$ together with a weight vector $\w=\{w_{uv} \mid uv\in E\}\cup \{b_{v} \mid v\in V\text{ is an internal neuron}\}$ defines a
predictor $h_{\cn,\w}:\cx\to\reals^k$ whose prediction
is given by ``propagating'' $\x$ forward through the network.
Concretely, we define $h_{v,\w}(\cdot)$ to be the output of the subgraph
of the neuron $v$ as follows: for an input neuron $v$, $h_{v,\w}$ outputs the corresponding coordinate in $\x$, and
internal neurons, we define $h_{v,\w}$ recursively as
$$h_{v,\w}(\x) = \sigma_v\left(\textstyle
	\sum_{u\in \IN(v)}\, w_{uv}\,h_{u,\w}(\x) + b_v\right)\,.$$
For output neurons, we define $h_{v,\w}$ as
$$h_{v,\w}(\x) = \textstyle
	\sum_{u\in \IN(v)}\, w_{uv}\,h_{u,\w}(\x)\,.$$
Finally, we let $h_{\cn,\w}(\x)=(h_{o_1,\w}(\x),\ldots,h_{o_k,\w}(\x))$.

We next describe the learning algorithm that we analyze in this paper. While there is no standard training algorithm for neural networks, the algorithms used in practice are usually quite similar to the one we describe, both in the way the weights are initialized and the way they are updated. 
We will use the popular Xavier initialization~\citep{glorot2010understanding} for the network weights. Fix $0\le \beta \le 1$.
We say that $\w^0=\{w^0_{uv}\}_{uv\in E}\cup \{b_{v}\}_{v\in V\text{ is an internal neuron}}$ are {\em $\beta$-biased random weights} (or, $\beta$-biased random initialization) if 
each weight
$w_{uv}$ is sampled independently from a normal distribution with mean $0$
and variance ${(1-\beta)d\delta(u)}/{\delta(\IN(v))}$ if $u$ is an input neuron and ${(1-\beta)\delta(u)}/{\delta(\IN(v))}$ otherwise. Finally, each bias term
$b_{v}$ is sampled independently from a normal distribution with mean $0$
and variance $\beta$.
We note that the rational behind this initialization scheme is that for every example $\x$ and every neuron $v$ we have $\E_{\w_0}\left(h_{v,\w_0}(\x)\right)^2=1$ (see \cite{glorot2010understanding})

\begin{algorithm}[ht]
	\caption{Generic Neural Network Training}
	\begin{algorithmic}\label{alg:general_nn_training}
		\STATE \textbf{Input: } Network $\cn$, learning rate $\eta>0$, batch size $m$, number of steps $T>0$, bias parameter $0\le \beta\le 1$, flag $\mathrm{zero\_prediction\_layer}\in \{\mathrm{True,False}\}$.
		\STATE Let $\w^0$ be $\beta$-biased random weights
		\IF {$\mathrm{zero\_prediction\_layer}$}
		\STATE Set $w^0_{uv}=0$ whenever $v$ is an output neuron
		\ENDIF
		\FOR {$t=1,\ldots,T$}
		\STATE Obtain a mini-batch $S_t=\{(\x^t_i,y^t_i)\}_{i=1}^m\sim\cd^m$
		\STATE Using back-propagation, calculate a stochastic gradient $\bv^t = \nabla \cl_{S_t}(\w^t)$
		\STATE Update $\w^{t+1} = \w^t - \eta \bv^t$
		\ENDFOR
	\end{algorithmic}
\end{algorithm}

\paragraph{Kernel classes.} A function $\kappa:\cx\times \cx\to \reals$ is
a {\em reproducing kernel}, or simply a kernel, if for every
$\x_1,\ldots,\x_r\in\cx$, the $r \by r$ matrix $\Gamma_{i,j} = \{\kappa(\x_i,\x_j)\}$
is positive semi-definite. Each kernel induces a Hilbert space
$\ch_{\kappa}$ of functions from $\cx$ to $\reals$ with a corresponding norm
$\|\cdot\|_{\kappa}$. For $\bh\in\ch_\kappa^k$ we denote $\|\bh\|_\kappa =\sqrt{\sum_{i=1}^k\|h_i\|^2_{\kappa}}$. A kernel and its corresponding space are {\em
normalized} if $\forall \x\in\cx,\;\kappa(\x,\x)=1$. 

Kernels give rise to popular benchmarks for learning algorithms. Fix a normalized kernel  $\kappa$ and $M>0$. 
It is well known that that for $L$-Lipschitz loss $\ell$, the SGD algorithm is guaranteed to return a function $\bh$ such that
$\E\cl_\cd(\bh) \le \min_{\bh'\in \ch^k_\kappa,\;\|\bh'\|_\kappa\le M}\cl_\cd(\bh')+\epsilon$
using $\left(\frac{LM}{\epsilon}\right)^2$ examples. 
In the context of multiclass classification, for $\gamma>0$ we define $\ell^{\gamma}:\reals^k\times [k]\to\reals$ by $\ell^\gamma(\hat y,y) = \ind [\hat y_y\le \gamma+\max_{y'\ne y}\hat y_{y'}]$.
We say that a distribution $\cd$ on $\cx\times [k]$ is $M$-separable w.r.t.\ $\kappa$ if there is $\bh^*\in\ch_\kappa^k$ such that $\|\bh^*\|_\kappa \le M$ and $\cl^1_\cd(\bh^*)=0$. In this case, the perceptron algorithm is guaranteed to return a function $\bh$ such that $\E\cl^{0-1}_\cd(\bh) \le \epsilon$ using $\frac{2M^2}{\epsilon}$ examples. We note that both for perceptron and SGD, the above mentioned results are best possible, in the sense that any algorithm with the same guarantees, will have to use at least the same number of examples, up to a constant factor.

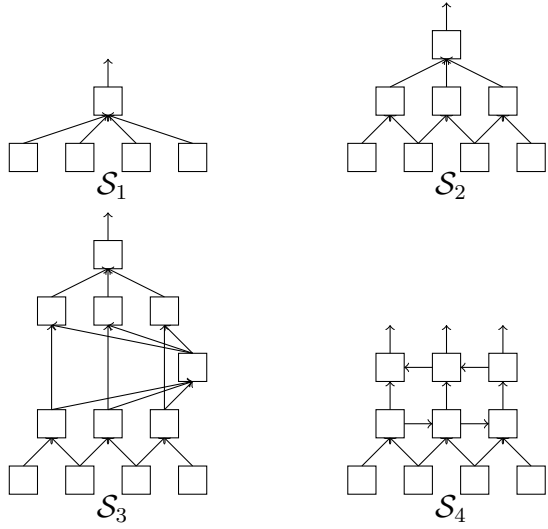
\begin{figure}[t]
\begin{center}
\begin{tikzpicture}[scale=0.75]
\foreach \i in {1,...,4}
{
	\draw (\i -7,0) rectangle (\i-7+0.5,0.5);
	\draw [->] (\i-7 +0.25  ,0.5) -- (-4.25 ,1);
}
\draw (-4.5,1) rectangle (-4,1.5);
\draw [->] (-4.25 ,1.5) -- (-4.25 ,2);
\node[text width=3cm] at (-2.45,-0.25) {$\cs_1$};
\foreach \i in {1,...,4}
{
	\draw (\i -1,0) rectangle (\i-1+0.5,0.5);
}
\foreach \i in {1,...,3}
{
	\draw (\i-1 + 0.5 ,1) rectangle (\i-1+1,1.5);
	\draw [->] (\i-1 +0.25  ,0.5) -- (\i-1 + 0.75 ,1);
	\draw [->] (\i-1 +1.25  ,0.5) -- (\i-1 + 0.75 ,1);
	\draw [->] (\i-1 + 0.75 ,1.5) -- (1.75 ,2);
}
\draw (1.5 ,2) rectangle (2,2.5);
\draw [->] (1.75 ,2.5) -- (1.75 ,3);
\node[text width=3cm] at (3.55,-0.25) {$\cs_2$};
\end{tikzpicture}
\begin{tikzpicture}[scale=0.75]
\foreach \i in {1,...,4}
{
	\draw (\i -7,0) rectangle (\i-7+0.5,0.5);
}
\foreach \i in {1,...,3}
{
	\draw (\i-7 + 0.5 ,1) rectangle (\i-7+1,1.5);
	\draw [->] (\i-7 +0.25  ,0.5) -- (\i-7 + 0.75 ,1);
	\draw [->] (\i-7 +1.25  ,0.5) -- (\i-7 + 0.75 ,1);
	\draw [->] (\i-7 + 0.75 ,1.5) -- (-2.75 ,2);
}
\draw (-3 ,2) rectangle (-2.5,2.5);
\foreach \i in {1,...,3}
{
	\draw (\i-7 + 0.5 ,3) rectangle (\i-7+1,3.5);
	\draw [->] (\i-7 +0.75, 1.5) -- (\i-7 +0.75,3);
	\draw [->] (-2.75, 2.5) -- (\i-7 +0.75,3);
	\draw [->] (\i-7 + 0.75 ,3.5) -- (-4.25 ,4);
}
\draw (-4.5 ,4) rectangle (-4,4.5);
\draw [->] (-4.25 ,4.5) -- (-4.25 ,5);
\node[text width=3cm] at (-2.45,-0.25) {$\cs_3$};
\foreach \i in {1,...,4}
{
	\draw (\i -1,0) rectangle (\i-1+0.5,0.5);
}
\foreach \i in {1,...,3}
{
	\draw (\i-1 + 0.5 ,1) rectangle (\i-1+1,1.5);
	\draw [->] (\i-1 +0.25  ,0.5) -- (\i-1 + 0.75 ,1);
	\draw [->] (\i-1 +1.25  ,0.5) -- (\i-1 + 0.75 ,1);
	\draw [->] (\i-1 + 0.75 ,1.5) -- (\i-1 + 0.75 ,2);
}
\draw [->] (1 ,1.25) -- (1.5 ,1.25);
\draw [->] (2 ,1.25) -- (2.5 ,1.25);
\foreach \i in {1,...,3}
{
	\draw (\i-1 + 0.5 ,2) rectangle (\i-1+1,2.5);
	\draw [->] (\i-1 + 0.75 ,2.5) -- (\i-1 + 0.75 ,3);
}
\draw [->] (1.5 ,2.25) -- (1 ,2.25);
\draw [->] (2.5 ,2.25) -- (2 ,2.25);
\node[text width=3cm] at (3.55,-0.25) {$\cs_4$};
\end{tikzpicture}
\caption{Examples of computation skeletons.\label{fig:cs_examples}}
\end{center}
\end{figure}

\paragraph{Computation skeletons~\citep{daniely2016toward}}
In this section we define a simple structure which we term a computation
skeleton. The purpose of a computational skeleton is to compactly describe
a feed-forward computation from an input to an output. A single skeleton encompasses a family of neural networks
that share the same skeletal structure. Likewise, it defines a
corresponding normalized kernel.
\begin{definition} A {\em computation skeleton} $\cs$ is a DAG with $n$ inputs, whose
non-input nodes are labeled by activations, and has a single output node $\out(\cs)$.
\end{definition}
Figure \ref{fig:cs_examples} shows four example skeletons, omitting
the designation of the activation functions. 
We denote by $|\cs|$ the number of non-input nodes of $\cs$. The following definition shows how a skeleton, accompanied with a
replication parameter $r\ge 1$ and a number of output nodes $k$,
induces a neural network architecture.

\begin{definition}[Realization of a skeleton]
Let $\cs$ be a computation skeleton and consider input coordinates in
$\sphere^{d-1}$ as in \eqref{eq:coordinates}. For $r, k \ge 1$ we
define the following neural network $\cn=\cn(\cs,r,k)$.
For each input node in $\cs$, $\cn$ has $d$ corresponding input
neurons with weight $1/d$. For each internal node $v\in \cs$ labelled by an activation
$\sigma$, $\cn$ has $r$ neurons $v^1,\ldots,v^r$, each with an
activation $\sigma$ and weight $1/r$. In addition, $\cn$ has $k$ output neurons
$o_1,\ldots,o_k$ with the identity activation $\sigma(x)=x$ and weight $1$.
There is an edge $v^iu^j\in E(\cn)$ whenever $uv\in
E(\cs)$.  For every output node $v$ in $\cs$, each neuron $v^j$ is
connected to all output neurons $o_1,\ldots,o_k$. We term $\cn$ the
{\em $(r,k)$-fold realization} of $\cs$. 
\end{definition}

\noindent
Note that the notion of the replication parameter $r$ corresponds, in
the terminology of convolutional networks, to the number of
channels taken in a convolutional layer and to the number of hidden neurons
taken in a fully-connected layer.

\begin{figure}[t]
\begin{center}
\begin{tikzpicture}
\foreach \i in {1,...,4}
{
	\draw (\i -1,0) rectangle (\i-1+0.5,0.5);
}
\foreach \i in {1,...,3}
{
	\draw (\i-1 + 0.5 ,1) rectangle (\i-1+1,1.5);
	\draw [->] (\i-1 +0.25  ,0.5) -- (\i-1 + 0.75 ,1);
	\draw [->] (\i-1 +1.25  ,0.5) -- (\i-1 + 0.75 ,1);
	\draw [->] (\i-1 + 0.75 ,1.5) -- (1.75 ,2);
}
\draw (1.5 ,2) rectangle (2,2.5);
\draw [->] (1.75 ,2.5) -- (1.75 ,3);
\draw (1.5 ,3) rectangle (2,3.5);
\draw [->] (1.75 ,3.5) -- (1.75 ,4);
\node[text width=3cm] at (3.15,-0.25) {$\cs$};
\draw [ultra thick, |->] (3.75,1.75) -- (4.75,1.75);
\foreach \i in {1,...,4}
{
	\draw (4.85+\i,0.25) circle [radius=0.1];
	\draw (5.15+\i,0.25) circle [radius=0.1];
}
\foreach \i in {-1,...,1}
{
	\foreach \j in {-2,...,2}
	{
		\draw (7.5+1.75*\i + 0.3*\j,1.25) circle [radius=0.1];
		\draw [->] (6.85+\i,0.35) -- (7.5+1.75*\i + 0.3*\j,1.15);
		\draw [->] (7.15+\i,0.35) -- (7.5+1.75*\i + 0.3*\j,1.15);

		\draw [->] (7.85+\i,0.35) -- (7.5+1.75*\i + 0.3*\j,1.15);
		\draw [->] (8.15+\i,0.35) -- (7.5+1.75*\i + 0.3*\j,1.15);
		\foreach \k in {-2,...,2}
		{
			\draw [->] (7.5+1.75*\i + 0.3*\j,1.35) -- (7.5+\k,2.15);
		}
	}
}
\foreach \j in {-2,...,2}
{
	\draw (7.5+\j,2.25) circle [radius=0.1];
	\foreach \k in {-2,...,2}
	{
		\draw[->] (7.5+\j,2.35) -- (7.5+\k,3.15);
	}
}
\foreach \j in {-2,...,2}
{
	\draw (7.5+\j,3.25) circle [radius=0.1];
	\draw[->] (7.5+\j,3.35) -- (6,3.9);
	\draw[->] (7.5+\j,3.35) -- (7,3.9);
	\draw[->] (7.5+\j,3.35) -- (8,3.9);
	\draw[->] (7.5+\j,3.35) -- (9,3.9);
}
\draw (6,4) circle [radius=0.1];
\draw (7,4) circle [radius=0.1];
\draw (8,4) circle [radius=0.1];
\draw (9,4) circle [radius=0.1];
\draw[->] (6,4.1) -- (6,4.3);
\draw[->] (7,4.1) -- (7,4.3);
\draw[->] (8,4.1) -- (8,4.3);
\draw[->] (9,4.1) -- (9,4.3);
\node[text width=3cm] at (8.1,-0.25) {$\cn(\cs,5,4)$};
\end{tikzpicture}
\end{center}
\caption{A $(5,4)$-realization of the computation
skeleton $\cs$ with $d=2$.\label{fig:ct_to_nn}}
\end{figure}
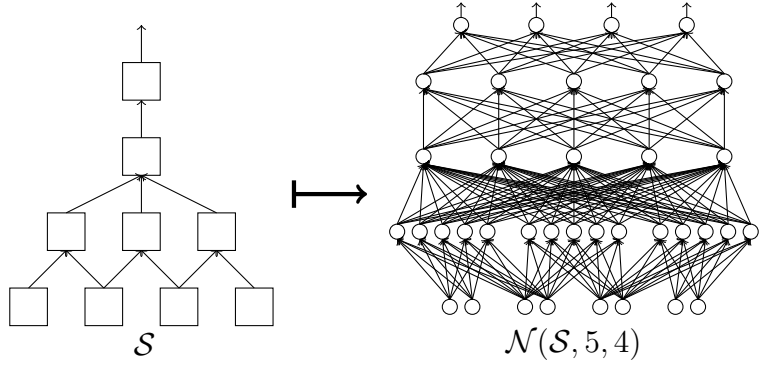

In addition to networks' architectures, a computation skeleton $\cs$ also
defines a normalized kernel $\kappa_\cs:\cx\times\cx\to[-1,1]$. To define the kernel,
we use the notion of a {\em conjugate activation}. For
$\rho\in[-1,1]$, we denote by $\gaussian_\rho$ the multivariate Gaussian
distribution on $\reals^2$ with mean $0$ and covariance matrix
$\left( \begin{smallmatrix} 1 & \rho \\ \rho & 1 \end{smallmatrix} \right)$.

\begin{definition}[Conjugate activation]\label{def:dual_act}
The {\em conjugate activation} of an activation $\sigma$ is the function
$\hat{\sigma}:[-1,1]\to\reals$ defined as
$
\hat\sigma(\rho) =
	\E_{(X,Y) \sim \gaussian_\rho}\sigma(X)\sigma(Y) \,.
$
\end{definition}
\noindent
The following definition gives the kernel corresponding to a skeleton
\begin{definition}[Compositional kernels]\label{def:comp_ker}
Let $\cs$ be a computation skeleton and let $0\le \beta\le 1$.
For every node $v$, inductively define a kernel
$\kappa^\beta_v:\cx\times\cx\to\reals$ as follows.
For an input node $v$ corresponding to the $i$th coordinate,
define $\kappa^\beta_{v}(\x,\y)=\inner{\x^i, \y^i}$.
For a non-input node $v$, define
$$
\kappa^\beta_v(\x,\y) =
	\hat\sigma_v\left(
		(1-\beta)\frac{\sum_{u\in \IN(v)}\kappa^\beta_{u}(\x,\y)}{|\IN(v)|}+\beta\right) \,.
    $$
The final kernel $\kappa^\beta_\cs$ is $\kappa^\beta_{\out(\cs)}$. The resulting Hilbert space and norm are denoted $\ch_{\cs,\beta}$ and $\|\cdot\|_{\cs,\beta}$ respectively.
\end{definition}

\section{Main results}

An activation $\sigma:\reals\to\reals$ is called {\em $C$-bounded} if $\|\sigma\|_\infty, \|\sigma'\|_\infty, \|\sigma''\|_\infty \le C$. Fix a skeleton $\cs$ and $1$-Lipschitz\footnote{If $\ell$ is $L$-Lipschitz, we can replace $\ell$ by $\frac{1}{L}\ell$ and the learning rate $\eta$ by $L\eta$. The operation of algorithm \ref{alg:general_nn_training} will be identical to its operation before the modification. Given this observation, it is very easy to derive results for general $L$ given our results. Hence, to save one paramater, we will assume that $L=1$.} convex loss $\ell$. 
Define $\comp(\cs)=\prod_{i=1}^{\depth(\cs)}\max_{v\in \cs, \depth(v)=i}(\deg(v)+1)$ and $\cc(\cs) = (8C)^{\depth(\cs)}\sqrt{\comp(\cs)}$, where $C$ is the minimal number for which all the activations in $\cs$ are $C$-bounded, and $\depth(v)$ is the maximal length of a path from an input node to $v$. We also define $\cc'(\cs) = (4C)^{\depth(\cs)}\sqrt{\comp(\cs)}$, where $C$ is the minimal number for which all the activations in $\cs$ are $C$-Lipschitz and satisfy $|\sigma(0)|\le C$. Through this and remaining sections we use $\gtrsim$ to hide universal constants. Likewise, we fix the bias parameter $\beta$ and therefore omit it from the relevant notation.

We note that for constant depth skeletons with maximal degree that is polynomial in $n$, $\cc(\cs)$ and $\cc'(\cs)$ are polynomial in $n$. These quantities are polynomial in $n$ also for various log-depth skeletons. For example, this is true for fully connected skeletons, or more generally, layered skeletons with constantly many layers that are not fully connected.

\begin{theorem}\label{thm:main_bounded_no_sep}
Suppose that all activations are $C$-bounded. Let $M,\epsilon>0$. Suppose that we run algorithm \ref{alg:general_nn_training} on the network $\cn(\cs,r,k)$ with the following parameters:
\begin{itemize}
\item $\eta=\frac{\eta'}{r}$ for $\eta'\lesssim \frac{\epsilon}{\left(\cc'(\cs)\right)^2}$
\item $T\gtrsim \frac{M^2}{\eta' \epsilon}$
\item $r\gtrsim \frac{C^4(T\eta')^2M^2\left(\cc'(\cs)\right)^4
\log\left(\frac{C|\cs|}{\epsilon\delta}\right)}{\epsilon^2} +d$
\item Zero initialized prediction layer
\item Arbitrary $m$
\end{itemize}
Then, w.p.\ $\ge 1-\delta$ over the choice of the initial weights, there is $t\in [T]$ such that $\E\cl_\cd(\w^t)\le \min_{\bh\in\ch^k_\cs,\;\|\bh\|_\cs\le M}\cl_\cd(\bh)+\epsilon$. Here, the expectation is over the training examples.
\end{theorem}

We next consider ReLU activations. Here, $\cc'(\cs) = (\sqrt{32})^{\depth(\cs)}\sqrt{\comp(\cs)}$. 
\begin{theorem}\label{thm:main_relu_no_sep}
Suppose that all activations are the ReLU. Let $M,\epsilon>0$. Suppose that we run algorithm \ref{alg:general_nn_training} on the network $\cn(\cs,r,k)$ with the following parameters:
\begin{itemize}
\item $\eta=\frac{\eta'}{r}$ for $\eta'\lesssim \frac{\epsilon}{\left(\cc'(\cs)\right)^2}$
\item $T\gtrsim \frac{M^2}{\eta' \epsilon}$
\item $r\gtrsim \frac{(T\eta')^2M^2\left(\cc'(\cs)\right)^4
\log\left(\frac{|\cs|}{\epsilon\delta}\right)}{\epsilon^2} +d$
\item Zero initialized prediction layer
\item Arbitrary $m$
\end{itemize}
Then, w.p.\ $\ge 1-\delta$ over the choice of the initial weights, there is $t\in [T]$ such that $\E\cl_\cd(\w^t)\le \min_{\bh\in\ch^k_\cs,\;\|\bh\|_\cs\le M}\cl_\cd(\bh)+\epsilon$. Here, the expectation is over the training examples.
\end{theorem}

Finally, we consider the case in which the last layer is also initialized randomly. Here, we provide guarantees in a more restricted setting of supervised learning. Concretely, we consider multiclass classification, when $\cd$ is separable with margin, and $\ell$ is the logistic loss.

\begin{theorem}\label{thm:main_bounded_sep}
Suppose that all activations are $C$-bounded, that $\cd$ is $M$-separable with w.r.t. $\kappa_\cs$ and let $\epsilon>0$. Suppose we run algorithm \ref{alg:general_nn_training} on $\cn(\cs,r,k)$ with the following parameters:
\begin{itemize}
\item $\eta=\frac{\eta'}{r}$ for $\eta'\lesssim \frac{\epsilon^2}{M^2 \left(\cc(\cs)\right)^4}$
\item $T\gtrsim \frac{\log(k)M^2}{\eta' \epsilon^2}$
\item $r\gtrsim C^4\left(\cc(\cs)\right)^4M^2\left(T\eta'\right)^2\log\left(\frac{C|S|}{\epsilon}\right) + k + d$ 
\item Randomly initialized prediction layer
\item Arbitrary $m$
\end{itemize}
Then, w.p.\ $\ge \frac{1}{4}$ over the choice of the initial weights and the training examples, there is $t\in [T]$ such that $\cl^{0-1}_\cd(\w^t)\le \epsilon$
\end{theorem}

\subsection{Implications}
To demonstrate our results, let us elaborate on a few implications for specific network architectures. To this end, let us fix the instance space $\cx$ to be either $\{\pm 1\}^n$ or $\sphere^{n-1}$. Also, fix a bias parameter $1\ge \beta>0$, a batch size $m$, and a skeleton $\cs$ that is a skeleton of a fully connected network of depth between $2$ and $\log(n)$. Finally, we also fix the activation function to be either the ReLU or a $C$-bounded activation, assume that the prediction layer is initialized to $0$, and fix the loss function to be some convex and Lipschitz loss function.
Very similar results are valid for convolutional networks with constantly many convolutional layers. We however omit the details for brevity.

Our first implication shows that SGD is guaranteed to efficiently learn constant degree polynomials with polynomially bounded weights. To this end, let us denote by $\cp_{t}$ the collection of degree $t$ polynomials. Furthermore, for any polynomial $p$ we denote by $\|p\|$ the $\ell^2$ norm of its coefficients.
\begin{corollary}\label{cor:polynomials}
Fix any positive integers $t_0,t_1$.
Suppose that we run algorithm \ref{alg:general_nn_training} on the network $\cn(\cs,r,1)$ with the following parameters:
\begin{itemize}
\item $\eta\lesssim \poly\left(\frac{\epsilon}{n}\right)$
\item $T,r\gtrsim \poly\left(\frac{n}{\epsilon},\log\left(1/\delta\right)\right)$
\end{itemize}
Then, w.p.\ $\ge 1-\delta$ over the choice of the initial weights, there is $t\in [T]$ such that $\E\cl_\cd(\w^t)\le \min_{p\in\cp_{t_0},\;\|p\|\le n^{t_1}}\cl_\cd(p)+\epsilon$. Here, the expectation is over the training examples.
\end{corollary}
We note that several hypothesis classes that were studied in PAC learning can be realized by polynomial threshold functions with polynomially bounded coefficients. This includes conjunctions, DNF and CNF formulas with constantly many terms, and DNF and CNF formulas with constantly many literals in each term.
If we take the loss function to be the logistic loss or the hinge loss, Corollary \ref{cor:polynomials} implies that SGD efficiently learns these hypothesis classes as well.

Our second implication shows that any continuous function is learnable (not necessarily in polynomial time) by SGD.
\begin{corollary}\label{cor:any_function}
Fix a continuous function $h^*:\sphere^{n-1}\to \reals$ and $\epsilon,\delta>0$. 
Assume that $\cd$ is realized\footnote{That is, if $(\x,y)\sim\cd$ then $y = h^*(\x)$ with probability $1$.} by $h^*$.
Assume that we run algorithm \ref{alg:general_nn_training} on the network $\cn(\cs,r,1)$. If $\eta>0$ is sufficiently small and $T$ and $r$ are sufficiently large,
then, w.p.\ $\ge 1-\delta$ over the choice of the initial weights, there is $t\in [T]$ such that $\E\cl_\cd(\w^t)\le \epsilon$.
\end{corollary}

\subsection{Extensions}
We next remark on two extensions of our main results. The extended results can be proved in a similar fashion to our results. To avoid cumbersome notation, we restrict the proofs to the main theorems as stated, and will elaborate on the extended results in an extended version of this manuscript.
First, we assume that the replication parameter is the same for all nodes. In practice, replication parameters for different nodes are different. This can be captured by a vector $\{r_v\}_{v\in Int(\cs)}$. Our main results can be extended to this case if for all $v$, $r_v\le \sum_{u\in\In(v)}r_{u}$ (a requirement that usually holds in practice).
Second, we assume that there is no weight sharing that is standard in convolutional networks. Our results can be extended to convolutional networks with weight sharing.

We also note that we assume that in each step of algorithm \ref{alg:general_nn_training}, a fresh batch of examples is given. In practice this is often not the case. Rather, the algorithm is given a training set of examples, and at each step it samples from that set. In this case, our results provide guarantees on the training loss. If the training set is large enough, this also implies guarantees on the population loss via standard sample complexity results.

\section{Proofs}
\paragraph{Notation} Throughout, we fix a loss $\ell:\reals^k\times \cy\to [0,\infty)$, a skeleton $\cs$, a replication parameter $r$, the network $\cn=\cn(\cs,r,k)$ and a bias parameter $0\le \beta\le 1$. For a matrix $W\in M_{r,l}(\reals)$ we denote $\|W\|_{p,q} = \max_{\|\x\|_p\le 1} \|W\x\|_q$, $\|W\|_2=\|W\|_{2,2}$, and $\|W\|_F=\sqrt{\sum_{i=1}^r\sum_{j=1}^l W_{ij}^2}$. We will often use the fact that $\|W\|_2\le \|W\|_F$. For $\sigma:\reals\to\reals$ and $\x\in \reals^n$ we abuse notation and denote $\sigma(\x)=(\sigma(x_1),\ldots,\sigma(x_n))$. 

For a skeleton $\cs$ we denote by $\Int(\cs)$ the set of $\cs$'s internal nodes. We will aggregate the weights of $\cn$ by a collection of matrices and bias vectors
\[
\W=\{W^v\}_{v\in \Int(\cs)}\cup \{\bb^v\}_{v\in \Int(\cs)}\cup\{W^\pred\}~.
\]
Here, $(W^v,\bb^v)$ are the matrix and vector that maps the output of all the neurons corresponding to nodes in $\In(v)$, to the neurons corresponding to $v$. 
Likewise, $W^\pred$ is the matrix that maps the output of the neurons corresponding to  $\out(\cs)$ to the final output of the network. We decompose $W^v$ further as a concatenation of two matrices $W^{v,\Int}, W^{v,\inp}$ that correspond to the internal and input nodes in $\In(v)$ respectively. 
For a prediction matrix $W^*\in M_{k,r}$ and weights $\W$ we denote by $\W|W^*$ the weights obtained by replacing $W^\pred$ with $W^*$.
We let
\[
\|\W\|_{2} = \max \left\{\|W^{v,\Int}\|_{2}, \frac{\|W^{v,\inp}\|_{2}}{\sqrt{r}} , \frac{\|\bb^v\|_{2}}{\sqrt{r}} : v\in\Int(\cs) \right\}\cup\{\|W^\pred\|_2\}~,
\]
\[
\|\W\|'_{2} = \max \left\{\|W^{v,\Int}\|_{2}, \frac{\|W^{v,\inp}\|_{2}}{\sqrt{r}} , \frac{\|\bb^v\|_{2}}{\sqrt{r}} : v\in\Int(\cs) \right\}~,
\]
and
\[
\|\W\|_F=\sqrt{\|W^\pred\|_F^2+\sum_{v\in \Int(\cs)}\|W^v\|^2_F + \|\bb^v\|^2_F}~.
\]
Finally, we let $\cw_{R} = \{\W : \|\W\|\le R\}$ and $\cw'_{R} = \{\W : \|\W\|'\le R\}$.
For $\x\in\cx$ we denote by $h_\W(\x)=R^\pred_\x(\W)$ the output on $\x$ of the network $\cn$ with the weights $\W$. Given $v\in V(\cs)$ we let $R^v_\x(\W)\in\reals^r$ to be the output of the neurons corresponding to $v$. We denote by $R_\x(\W):=R^{\out(\cs)}_\x(\W)$ the output of the representation layer. 
We also let $R^{v\leftarrow}_\x(\W)$ be the concatenation of $\{R^u_\x(\W)\}_{u\in \In(v)}$. Note that $R^{v}_\x(\W) = \sigma_v(W^vR^{v\leftarrow}_\x(\W))$ and $R^{\pred}_\x(\W) = W^\pred R^{\out(\cs)}_\x(\W)$. For $(\x,y)\in\cx\times\cy$ we denote $\ell_{(\x,y)}(\W)=\ell(R^\pred_\x(W),y)$ and for $S=\{(\x_i,y_i)\}_{i=1}^m$ we denote $\cl_S(\W)=\frac{1}{m}\sum_{i=1}^m\ell_{(\x,y)}(\W)$. We let $\cl_\cd(\W)=\E_{(\x,y)\sim\cd}\ell_{(\x,y)}(\W)$. Finally, we let $k_\W(\x,\x') = \frac{\inner{R_\x(\W),R_{\x'}(\W)}}{r}$.

\subsection{Overview}
We next review the proof of theorem \ref{thm:main_relu_no_sep}. The proof of theorem \ref{thm:main_bounded_no_sep} is similar. Later, we will also comment how the proof can be modified to establish theorem \ref{thm:main_bounded_sep}.
Let $\bh^*\in \ch^k_\cs$ be some function with $\|\bh^*\|\le M$ and let $\W_0,\ldots,\W_T$ be the weights produced by the SGD algorithm. Our goal is to show that w.h.p.\ over the choice of $\W_0$, there is $t\in [T]$ such that $\E\cl_\cd(\W_t)\le \epsilon$.

In section \ref{sec:init} we show that w.h.p.\ over the choice of $\W_0$, there is a prediction matrix $W\in M_{k,r}$ so that $\cl_\cs(\W_0|W^*)\le \epsilon$ and $\|W^*\|_F\le\frac{M}{\sqrt{r}}$. This follows from the results of \cite{daniely2016toward}, and some extensions of those. Namely, we extend the original from $k=1$ to general $k$, and also eliminate a certain logarithmic dependence on the size of the support of $\cd$.

Given that such $W^*$ exists, standard online learning results (e.g.\ Chapter 21 in
\cite{shalev2014understanding}) imply that if we would apply SGD only on the last layer, with the learning rate specified in theorem \ref{thm:main_relu_no_sep}, i.e.\, $\eta=\frac{\eta'}{r}$ for $\eta'\lesssim \frac{\epsilon}{\left(\cc'(\cs)\right)^2}$, we would be guaranteed to have some step $t\in [T]$ in which $\E\cl_\cd(\W_t)\le 2\epsilon$.

However, as we consider SGD on all weights, this is not enough. Hence, in section \ref{sec:opt}, we show that with the above mentioned learning rate, the weights of the non-last layer change slowly enough, so that $\cl_\cs(\W_t|W^*)\le \epsilon$ for all $t$. Given this, we can invoke the online-learning based argument again.

In order to show that the last layer changes slowly, we need to bound the magnitude of the gradient of the training objective. In section \ref{sec:boundness} we establish such a bound on the gradient of the loss for every example. As $\cl_\cd(\W)$ and $\cl_{S_t}(\W)$ are averages of such functions, the same bound holds for them as well. We note that our bound depends on the spectral norm on the matrices $\{W^v\}_{v\in\cs}$. We show that for random matrices, w.h.p.\, the magnitude of the norm implies a bound that is good enough for our purposes. Likewise, trough the training process, the norm doesn't grow too much, so the desired bound is valid throughout the optimization process.

The structure of the proof of theorem \ref{thm:main_bounded_sep} is similar, but has a few differences. First, the first step would be to show that in the case that $\cd$ is $M$-separable w.r.t.\ $\kappa_\cs$, then w.h.p. over the choice of $\W_0$, there is a prediction matrix $W^*\in M_{k,n}$ such that $\cl^1_\cs(\W_0|W^*)$ is tiny, and $\|W^*\|_F\lesssim\frac{M}{\sqrt{r}}$. Again, this is based on the results and techniques of \cite{daniely2016toward}, and is done in section \ref{sec:init}. Given this, again, running SGD on the top layer would be fine. However, now we cannot utilize the online-learning based argument we used before, because the starting point is not $0$, but rather a random vector, whose norm is too large to carry out the analysis. In light of that, we take a somewhat different approach. 

We show that the weights beneath the last layer are changing slow enough, so that the following holds throughout the optimization process: As long as the 0-1 error is larger than $\epsilon$, the magnitude of the gradient is $\Omega\left(\frac{\epsilon\sqrt{r}}{M}\right)$. More precisely, the derivative in the direction of $W^*$, is smaller than $-\Omega\left(\frac{\epsilon\sqrt{r}}{M}\right)$. Given this, and bounds on both the first and second derivative of the loss (proved in section \ref{sec:boundness}), we are able to establish the proof by adopting a standard argument from smooth convex optimization (done in section \ref{sec:opt}).

\subsection{Boundness of the objective function}\label{sec:boundness}
Let $\Omega\subset \reals^n$ be an open set. For a function $\f:\Omega\to \reals^m$, a unit vector $\bu\in \reals^n$ and $\x_0\in \Omega$ we denote $\f_{\x_0,\bu}(t)=\f(\x_0 + t\bu)$. We say that $\f$ is $(\alpha,\beta,\gamma)$-bounded at $\x_0$ if $f$ is twice differentiable and
\[
\forall \bu\in \sphere^{n-1},\;\; \|\f_{\x_0,\bu}(0)\| \le \alpha,\;\; \|\f'_{\x_0,\bu}(0)\| \le \beta ,\;\; \|\f''_{\x_0,\bu}(0)\| \le \gamma ~.
\]
We say that $\f$ is $(\alpha,\beta,\gamma)$-bounded if it is $(\alpha,\beta,\gamma)$-bounded in any $\x_0\in \Omega$.
We note that for $m=1$, $f$ is $(\alpha,\beta,\gamma)$-bounded at $\x_0$ if and only if $|f(\x_0)|\le \alpha$, $\|\nabla f(\x_0)\|\le \beta$ and $\|\nabla^2f(\x_0)\|_{2}\le \gamma$. In particular, when $n=1$ too, $f$ is $(\alpha,\beta,\gamma)$-bounded at $\x_0$ if and only if $|f(\x_0)|\le \alpha$, $ |f'(\x_0)|\le \beta$ and $|f''(\x_0)|\le \gamma$. We will say that $\f$ is $C$-bounded if it is $(C,C,C)$-bounded.

\begin{fact}\label{fact:compose_with_act}
Let $\sigma:\reals\to\reals$ be $(\alpha,C,C)$-bounded function. Suppose that $\f:\Omega\to\reals^n$ is $(\infty,\beta,\beta^2)$-bounded. We have that $\g= \sigma\circ \f$ is $(\infty,C\beta,2C\beta^2)$-bounded. If we furthermore assume that $\sigma(0)=0$ then we have that $\g$ is $(C\alpha,C\beta,2C\beta^2)$-bounded. \end{fact}
\proof
The first part follows from the facts that
\[
\g_{\x_0,\bu}'(t) = \sigma'(\f_{\x,\bu}(t)) \f_{\x,\bu}'(t)
\]
and
\[
\g_{\x_0,\bu}''(t) = \sigma''(\f_{\x,\bu}(t)) \left(\f_{\x,\bu}'(t)\right)^2 +  \sigma'(\f_{\x,\bu}(t)) \f_{\x,\bu}''(t)
\]
The second part follows from the fact that in the case that $\sigma(0)=0$ we have that $\|\sigma(\x)\|\le C\|x\|$.
\proofbox

\begin{fact}\label{fact:compose_with_loss}
Let $l:\reals^n\to\reals$ be $(\infty,C,C)$-bounded function. Suppose that $\f:\Omega\to\reals^n$ is $(\infty,\beta,\beta^2)$-bounded. We have that $g= l\circ \f$ is $(\infty,C\beta,2C\beta^2)$-bounded
\end{fact}
\proof
This follows from the fact that
\[
g_{\x_0,\bu}'(t) = \inner{\nabla l(\f_{\x,\bu}(t)), \f_{\x,\bu}'(t)}
\]
and
\[
g_{\x_0,\bu}''(t) = \inner{\f_{\x,\bu}'(t), H_l(\f_{\x,\bu}(t)) \f_{\x,\bu}'(t)} +   \inner{\nabla l(\f_{\x,\bu}(t)), \f_{\x,\bu}''(t)}
\]
\proofbox

\begin{example}[logistic loss]\label{exam:log_loss}
Recall that $\p:\reals^k\to \Delta^{k-1}$ is given by $p_i(\hat y) = \frac{e^{\hat y_i}}{\sum_{j=1}^ke^{\hat y_{j}}}$ and $\ell:\reals^k\times\cy\to\reals_+$ by $\ell(\hat y,y)= -\log\left(p_y(\hat y)\right)$. Denote $\ell_y(\hat y)=\ell(\hat y,y)$.
We have
\[
\frac{\partial p_i}{\partial \hat y_j} = \frac{\delta_{ij}e^{\hat y_i}\left(\sum_{j=1}^ke^{\hat y_j}\right) - 
e^{\hat y_i}e^{\hat y_j}
}{\left(\sum_{j=1}^ke^{\hat y_j}\right)^2}
\]
Hence, $\nabla p_i = p_i\e_i - p_i\p$ and therefore $\nabla \ell_i = -\frac{\nabla p_i}{p_i} = -\e_i + \p$. Hence, $\nabla^2 \ell_i = \nabla \p = \diag(\p) - \p\otimes\p$. In particular, $\ell_y$ is $(\infty,\sqrt{2},1)$-bounded.
\end{example}

\begin{fact}\label{fact:compose_with_lin}
Let $B$ be the set of $l\times m$ matrices with operator norm less than $R$ and let $\f:B\times\Omega\to\reals^m$. Define $\g:B\times \Omega\to \reals^l$ by $\g(W,\x) = W\f(\x,W)$. Then, if $\f$ is $(\alpha,\beta,\gamma)$-bounded then $\g$ is $(R\alpha,R\beta + \alpha,R\gamma + 2\beta)$-bounded
\end{fact}
\proof
Fix $(W_0,\x_0)\in B\times \Omega$ and $(U,\bu)\in M_{l,m}(\reals)\times \reals^{n}$ such that $\|U\|^2_F + \|u\|^2_2 =1$.
We have
\[
\g_{(W_0,\x_0),(U,\bu)}(t) = W_0\f_{(W_0,\x_0),(U,\bu)}(t) + tU\f_{(W_0,\x_0),(U,\bu)}(t)
\]
Hence, 
\[
\g'_{(W_0,\x_0),(U,\bu)}(t) = W_0\f'_{(W_0,\x_0),(U,\bu)}(t) + U\f_{(W_0,\x_0),(U,\bu)}(t) + tU\f'_{(W_0,\x_0),(U,\bu)}(t)
\]
\[
\g''_{(W_0,\x_0),(U,\bu)}(t) = W_0\f''_{(W_0,\x_0),(U,\bu)}(t) + 2U\f'_{(W_0,\x_0),(U,\bu)}(t) + tU\f''_{(W_0,\x_0),(U,\bu)}(t)
\]
\proofbox

\begin{fact}\label{fact:concat}
If $\f_1,\ldots,\f_d:\Omega\to \reals^m$ are $(\alpha,\beta,\gamma)$-bounded then $(\f_1,\ldots,\f_d):\Omega\to \reals^{mk}$ is $(\sqrt{d}\alpha,\sqrt{d}\beta,\sqrt{d}\gamma)$-bounded
\end{fact}

From facts \ref{fact:compose_with_act}, \ref{fact:compose_with_lin} and \ref{fact:concat} we conclude that

\begin{lemma}\label{lem:adding_node}
Suppose that
\begin{itemize}
\item $\f_{1,1},\ldots,\f_{1,d_1}:\Omega\to \reals^r$ are $(\alpha,\beta,\beta^2)$-bounded functions with $\sqrt{r}\le \alpha\le\beta$
\item $\f_{2,1},\ldots,\f_{2,d_2}:\Omega\to \reals^d$ are $(1,0,0)$-bounded functions
\item $\sigma:\reals\to\reals$ is $C$-bounded  for $C\ge1$.
\item Let $B_1$ be the set of $r\times rd_1$ matrices with operator norm less than $R$ for $R\ge 1$, $B_2$ be the set of $r\times dd_2$ matrices with operator norm less than $r\sqrt{r}$ for $R\ge 1$, and $B_3\subset \reals^r$ the set of vectors with norm less than $R\sqrt{r}$.
Define $\g:B_1\times B_2\times B_3\times \Omega\to \reals^r$ by $\g(W^1,W^2,\bb,\x) = \sigma(W^1\f_1(\x) + W^2\f_2(\x) + \bb)$ where $\f_i(\x)= (\f_{i,1}(\x),\ldots,\f_{i,d_i}(\x))$
\end{itemize}
Then, $\g$ is  $(C\sqrt{r},  4CR\sqrt{\tilde d}\beta ,  (4CR\sqrt{\tilde d}\beta)^2)$-bounded for $\tilde d = d_1+d_2+1$.
\end{lemma} 
\proof
By fact \ref{fact:concat}, $\f_1$ is $(\sqrt{d_1}\alpha,\sqrt{d_1}\beta,d\beta^2)$-bounded 
Hence, by fact \ref{fact:compose_with_lin}, $(W^1,\x)\mapsto W^1\f_1(\x)$ is $(R\sqrt{d_1}\alpha,  R\sqrt{d_1}\beta + \sqrt{d_1}\alpha ,  Rd_1\beta^2 + 2\sqrt{d_1}\beta)$-bounded. Since $\beta\ge \max(1,\alpha)$ and $R\ge 1$ we have that $(W^1,\x)\mapsto W^1\f_1(\x)$ is $(R\sqrt{d_1}\alpha,  2R\sqrt{d_1}\beta ,  (2R\sqrt{d_1}\beta)^2)$-bounded. Similarly, $(W^2,\x)\mapsto W^2\f_2(\x)$ is $(R\sqrt{d_2r},\sqrt{d_2},0)$-bounded and $(\bb,\x)\mapsto \bb$ is $(R\sqrt{r},1,0)$-bounded. As $\sqrt{r}\le \alpha$, and $\beta,R\ge 1$ it follows that
$(W^1,W^2,\bb,\x)\mapsto W^1\f_1(\x) + W^2\f_2(\x) + \bb$ is $(R\sqrt{\tilde d}\alpha,  2R\sqrt{\tilde d}\beta ,  (2R\sqrt{\tilde d}\beta)^2)$-bounded.

Now, by fact  \ref{fact:compose_with_act} and the $C$-boundness of $\sigma$, $\g$ is $(C\sqrt{r},  2CR\sqrt{\tilde d}\beta ,  2C(2R\sqrt{\tilde d}\beta)^2)$-bounded. The lemma concludes as $C\ge 1$
\proofbox

\begin{lemma}\label{lem:adding_node_unblounded}
Suppose that
\begin{itemize}
\item $\f_{1,1},\ldots,\f_{1,d_1}:\Omega\to \reals^r$ are $(\beta,\beta,\infty)$-bounded functions with $\sqrt{r}\le \beta$
\item $\f_{2,1},\ldots,\f_{2,d_2}:\Omega\to \reals^d$ are $(1,0,0)$-bounded functions
\item $\sigma:\reals\to\reals$ is $C$-Lipschitz and satisfy $|\sigma(0)|\le C$.
\item Let $B_1$ be the set of $r\times rd_1$ matrices with operator norm less than $R$ for $R\ge 1$, $B_2$ be the set of $r\times dd_2$ matrices with operator norm less than $r\sqrt{r}$ for $R\ge 1$, and $B_3\subset \reals^r$ the set of vectors with norm less than $R\sqrt{r}$.
Define $\g:B_1\times B_2\times B_3\times \Omega\to \reals^r$ by $\g(W^1,W^2,\bb,\x) = \sigma(W^1\f_1(\x) + W^2\f_2(\x) + \bb)$ where $\f_i(\x)= (\f_{i,1}(\x),\ldots,\f_{i,d_i}(\x))$
\end{itemize}
Then, $\g$ is  $(C(R+1)\sqrt{\tilde d}\beta,  C(R+1)\sqrt{\tilde d}\beta ,  \infty)$-bounded for $\tilde d = d_1+d_2+1$.
\end{lemma} 
\proof
By fact \ref{fact:concat}, $\f_1$ is $(\sqrt{d_1}\beta,\sqrt{d_1}\beta,\infty)$-bounded 
Hence, by fact \ref{fact:compose_with_lin}, $(W^1,\x)\mapsto W^1\f_1(\x)$ is $(R\sqrt{d_1}\beta,  R\sqrt{d_1}\beta + \sqrt{d_1}\beta ,  \infty)$-bounded. Similarly, $(W^2,\x)\mapsto W^2\f_2(\x)$ is $(R\sqrt{d_2r},\sqrt{d_2},0)$-bounded and $(\bb,\x)\mapsto \bb$ is $(R\sqrt{r},1,0)$-bounded. As $\sqrt{r}\le \beta$ and $\beta,R\ge 1$ it follows that
$(W^1,W^2,\bb,\x)\mapsto W^1\f_1(\x) + W^2\f_2(\x) + \bb$ is $((R+1)\sqrt{\tilde d}\beta,  (R+1){\tilde d}\beta ,\infty)$-bounded.
The lemma concludes by fact  \ref{fact:compose_with_act} and the $C$-boundness of $\sigma$.
\proofbox

Using a similar argument one can prove that
\begin{lemma}\label{lem:adding_loss}
Suppose that
\begin{itemize}
\item $\f:\Omega\to \reals^r$ is $(\alpha,\beta,\beta^2)$-bounded function with $\max(1,\alpha)\le\beta$
\item $l:\reals^k\to\reals$ is $(\infty,C,C)$-bounded for $C\ge 1$.
\item Let $B$ be the set of $k\times r$ matrices with operator norm less than $R$ for $R\ge 1$. Define $g:B\times \Omega\to \reals$ by $g(W,\x) = l(W\f(\x))$
\end{itemize}
Then, $g$ is  $(\infty,  4CR\beta ,  (4CR\beta)^2)$-bounded
\end{lemma}

\begin{lemma}\label{lem:boundness}
Assume that all activations in $\cs$ are $C$-bounded and that each $\ell_y$ is $(\infty,C',C')$-bounded.
Let $R\ge 1$ and $(\x,y)\in\cx\times\cy$. Then
\begin{itemize}
\item The function $\ell_{(\x,y)}:\cw_R \to \reals$ is $(\infty,\beta,\beta^2)$-bounded for $\beta=4C'R(4CR)^{\depth(\cs)}\sqrt{\comp(\cs)r}$.
\item The function $R_\x:\cw_R\to\reals^r$ is $(C\sqrt{r},\beta,\beta^2)$-bounded for $\beta=(4CR)^{\depth(\cs)}\sqrt{\comp(\cs)r}$.
\end{itemize}
\end{lemma}

\proof
Denote $\comp^i(\cs) = \prod_{j=1}^i\max_{v\in\cs,\depth(v)=j} (\deg(v)+1)$ and
$\beta_i = (4CR)^i\sqrt{r\cdot \comp^i(\cs)}$. We will prove that for every $v\in \cs$, the function $R^v_\x(\W)$ is $(C\sqrt{r},\beta_{\depth(v)},\beta_{\depth(v)}^2)$-bounded.
This proves the second item. The first item follows from the second together with lemma \ref{lem:adding_loss}.

We will use induction on $\depth(v)$.
For depth $0$ node (i.e., an input node), the function $R^v_{\x}$ is a constant function with output of norm $1$. Hence, it is $(1,0,0)$-bounded. 
For $v$ of depth $>0$, the induction hypothesis and lemma \ref{lem:adding_node} implies that $R^v_\x$ is
\[
(C\sqrt{r},4CR\sqrt{\deg(v)+1}\beta_{i-1},(4CR\sqrt{\deg(v)+1}\beta_{i-1})^2)\text{-bounded}
\]
The proof concludes as $\beta_i\le 4CR\sqrt{\deg(v)+1}\beta_{i-1}$
\proofbox

Based on lemma \ref{lem:adding_node_unblounded} and a similar argument we have that:
\begin{lemma}\label{lem:boundness_relu}
Assume that all activations in $\cs$ are $C$-Lipschitz and satisfy $|\sigma(0)|\le C$, and that $\ell$ is $L$-Lipschitz.
Let $R\ge 1$ and $(\x,y)\in\cx\times\cy$. 
Let $\beta = (C(R+1))^{\depth(\cs)}\sqrt{\comp(\cs)r}$.
Then,
\begin{itemize}
\item The function $\ell_{(\x,y)}:\cw_{R} \to \reals$ is 
$(\infty,LR\beta,\infty)$-bounded when restricted to the variables $\W^{\inter} = \{W^v\}_{v\in\cs}$
\item The function $\ell_{(\x,y)}:\cw_{R} \to \reals$ is 
$(\infty,L\beta,\infty)$-bounded when restricted to the variables $W^{\pred}$
\item The function $R_\x$ is $(\beta,\beta,\infty)$-bounded
\end{itemize}
If we furthermore assume that all activations satisfy $\|\sigma\|_\infty\le C$ then (i) in the last item the conclusion is that $R_\x:\cw_{R} \to \reals$ is $(C\sqrt{r},\beta,\infty)$-bounded, and (ii) $\ell_{(\x,y)}$ is 
$(\infty,LC\sqrt{r},\infty)$-bounded when restricted to the variables $W^{\pred}$
\end{lemma}

\subsection{Optimization given good initialization}\label{sec:opt}

\begin{lemma}\label{lem:gd_change}
Assume that all activations in $\cs$ are $C$-Lipschitz and satisfy $|\sigma(0)|\le C$, and that $\ell$ is $L$-Lipschitz. Define  $\alpha = 2L(3C)^{\depth(\cs)}\sqrt{\comp(\cs)}$.
Let $\W_0$ be initial weights with $\W_0\in\cw_{1.5}$. Suppose that $\W_1,\ldots,\W_t$ are the weights obtained by running algorithm \ref{alg:general_nn_training} with learning rate $\eta' = \frac{\eta}{r}>0$. Then, for every $t\le \frac{\sqrt{r}}{2\eta \alpha}$ we have
\begin{enumerate}
\item\label{item:1} $\W_t\in\cw_{2}$
\item\label{item:3} For any and every $\x$, $\|R_\x(\W_0) - R_\x(\W_t)\|\le t\eta\alpha^2$
\end{enumerate}
\end{lemma}

\proof
Let $\V_t$ the stochastic gradient at time $t$.
To see item \ref{item:1} note that as long as $\W_t \in \cw_2$, by lemma \ref{lem:boundness_relu},  for all $v\in V(\cs)\cup\{\pred\}$, $\|V^v_t\|_{F}\le \alpha\sqrt{r}$. Since $\|V^v_t\|_2\le \|V^v_t\|_F$, and since the learning rate is $\frac{\eta}{r}$, at each step, the spectral norm of each $W^v$ is changed by at most $\eta\alpha\sqrt{\frac{1}{r}}$. Hence, as long at $t\le  \frac{\sqrt{r}}{2\eta\alpha}$, we have that  $W_t\in \cw_{2}$.
For item \ref{item:3}, again since  $\|\V^{\inter}_t\|_{F}\le \alpha\sqrt{r}$, we have that the euclidian length of the trajectory of the internal weights until step $t$ is at most $\frac{\eta t\alpha}{\sqrt{r}}$. Now, by lemma \ref{lem:boundness_relu}, $R_\x(\W)$ is $(\alpha\sqrt{r})$-Lipschitz. Hence, $\|R_\x(\W_0) - R_\x(\W_t)\|\le t\eta\alpha^2$ for all $\x$. 
\proofbox

\subsubsection{Starting from zero prediction layer}
We will use the following fact from online convex optimization.
\begin{theorem}[e.g.\ Chapter 21 in
\cite{shalev2014understanding}]\label{thm:oco}
Let $f_1,\ldots,f_T:\reals^n\to \reals$ be $L$-Lipschitz convex functions. Let $\x_0 = 0$ and $\x_{t+1} = \x_t - \eta\nabla f_t(\x_{t})$. Here, $\nabla f_t(\x_{t})$ is some sub-gradient of $f_t$ at $\x_t$. Then, for any $\x^*\in\reals^n$ we have,
\[
\sum_{t=1}^T f_t(\x_t) \le \sum_{t=1}^T f_t(\x^*) + \frac{\|\x^*\|^2}{2\eta} + \frac{\eta TL^2}{2}
\]
\end{theorem}

\begin{lemma}\label{lem:guarantee_given_good_init_relu}
Assume that all activations in $\cs$ are $C$-Lipschitz and satisfy $|\sigma(0)|\le C$, and that $\ell$ is $L$-Lipschitz. Define  $\alpha = 2L(3C)^{\depth(\cs)}\sqrt{\comp(\cs)}$. Fix $\epsilon >0$, a prediction matrix $W^*\in M_{k,r}$ with $\|W^*\|_F\le \frac{M}{\sqrt{r}}$, and $\W_0\in \cw_{1.5}$ with $\W_0^\pred=0$.
Let $\W_1,\ldots,\W_T$ be the weights produced by algorithm \ref{alg:general_nn_training} with step size $\eta' = \frac{\eta}{r}$ for $\eta \le \frac{8\epsilon}{\alpha^2}$, with $T\ge \frac{M^2}{2\eta\epsilon}$ and with arbitrary batch size $m$. Furthermore, assume that 
\[
\sqrt{r}\ge \frac{MLT\eta\alpha^2}{\epsilon} + 2T\eta\alpha
\]
Then, there is some $t\in [T]$ such that $\E\cl_{\cd}(\W_t)\le \cl_{\cd}(\W_0|W^*) +3\epsilon$.

\end{lemma}
\proof 
By Lemma \ref{lem:gd_change} we have that for all $t$, 
\[
\cl_{\cd}(\W_t|W^*) = \E_{(\x,y)\sim\cd}\ell (W^*R_\x(\W_t),y)\le \cl_{\cd}(\W_0|W^*) + \frac{MLt\eta\alpha^2}{\sqrt{r}}
\]
Since $\sqrt{r}\ge \frac{MLT\eta\alpha^2}{\epsilon}$ we have that
\begin{equation}\label{eq:good_loss}
\cl_{\cd}(\W_t|W^*)\le \cl_{\cd}(\W_0|W^*)+\epsilon
\end{equation}
throughout the optimization process. Likewise, $\W_t\in \cw_2$ for all $t\in [T]$, and therefore $\|R_\x(\W_t)\|\le \frac{\alpha\sqrt{r}}{2L}$.
Now, consider the convex functions $f_t:M_{k,r}\to\reals$ defined by $f_t(W)= L_{S_t}(\W_0|W)$. Since $\|R_\x(\W_t)\|\le \frac{\alpha\sqrt{r}}{2L}$ and $\ell$ is $L$-Lipschitz, $f_t$ is $\left(\frac{\alpha\sqrt{r}}{2}\right)$-Lipschitz.
Hence, applying theorem \ref{thm:oco} we conclude that
\begin{eqnarray*}
\sum_{t=1}^T \cl_{S_t}(\W_t) &\le& \sum_{t=1}^T \cl_{S_t}(\W_t | W^*)  + \frac{\frac{M^2}{r}}{2\eta'} + \frac{T\eta'\alpha^2r}{8}
\\
&=& \sum_{t=1}^T \cl_{S_t}(\W_t | W^*)  + \frac{M^2}{2\eta} + \frac{T\eta\alpha^2}{8}
\end{eqnarray*}
Now, since $T\ge \frac{M^2}{2\eta\epsilon}$ and $\eta\le \frac{\epsilon 8 }{\alpha^2}$
we have that
\[
\frac{1}{T}\sum_{t=1}^T \cl_{S_t}(\W_t) \le \frac{1}{T}\sum_{t=1}^T \cl_{S_t}(\W_t | W^*)  +2\epsilon
\]
Taking expectation (w.r.t.\ the mini-batches) and using equation (\ref{eq:good_loss}) we get 
\[
\frac{1}{T}\sum_{t=1}^T \E\cl_{\cd}(\W_t) \le \cl_{\cd}(\W_0|W^*)+3\epsilon
\]
In particular, there is some $t\in [T]$ for which $\E\cl_{\cd}(\W_t)\le \cl_{\cd}(\W_0|W^*)+3\epsilon$
\proofbox

\subsubsection{Starting from random prediction layer}
In this section we assume that $\cy = [k]$, that $\ell$ is the logistic-loss, and that all acitivations are $C$-bounded. Denote $\alpha=8\sqrt{2}(8C)^{\depth(\cs)}\sqrt{\comp(\cs)}$

\begin{claim}\label{claim:improvement_dir}
Let $\cd$ be a distribution on $\reals^r\times [k]$ such that $\Pr_{(\x,j)\sim\cd}\left(\|x\|> C\sqrt{r}\right) = 0$.
Let $W^*\in M_{k\times r}$ be a matrix such that $\cl_\cd^1(W)=\E_{(\x,y)\sim\cd}\ell^1(W\x,y)\le \delta$. Then, for all $W\in M_{k\times r}$
\[
\|\nabla \cl_{\cd}(W)\| \ge \frac{\cl^{0-1}_\cd(W)-\delta}{2\|W^*\|_F} - \delta\sqrt{2r}C
\]
\end{claim}
\proof
Let $E=\frac{W^*}{\|W^*\|}$. Fix an example $(\x,y)$ and denote $\ell_{(\x,y)}(t)=\ell_y((W+tE)\x)$. We have that $\|\nabla \cl_{\cd}(W)\| \ge \inner{\nabla \cl_{\cd}(W),-E} = -\E_{(\x,y)\sim\cd}\ell_{(\x,y)}'(0)$.
By example \ref{exam:log_loss} we have
\[
-\ell_{(\x,y)}'(0)=\inner{E\x,\e_y - \p(W\x)} = \sum_{i=1}^k p_i(W\x)\left[(E\x)_y-(E\x)_i\right] = \sum_{i\ne y} p_i(W\x)\left[(E\x)_y-(E\x)_i\right]
\]
Now, if $\ell^1(W^*\x,y)=0$ then $(E\x)_y-(E\x)_i\ge \frac{1}{\|W^*\|_F}$ for all $i\ne y$ and hence $\ell_{(\x,y)}'(0)\ge \frac{1}{\|W^*\|_F}\sum_{i\ne y} p_i(W\x) = \frac{1-p_y(W\x)}{\|W^*\|_F}\ge 0$. If furthermore $\ell^{0-1}(W\x,y)=1$ then $p_y(W\x)\le \frac{1}{2}$. In this case, $-\ell_{(\x,y)}'(0)\ge\frac{1}{2\|W^*\|_F}$. On the other hand, we always have $\ell_{(\x,y)}'(0)\le \|E\|\cdot \|\x\|\cdot \|\e_y-\p(W\x)\|\le C\sqrt{2r}$. It follows that
\begin{eqnarray*}
\|\nabla \cl_{\cd}(W)\| &\ge& \Pr_{(\x,y)\sim\cd}\left(\ell^{0-1}_{y}(W\x)=1\text{ and }\ell^{1}_{y}(W^*\x)=0\right)\frac{1}{2\|W^*\|_F} -
\Pr_{(\x,y)\sim\cd}\left(\ell^{1}_{y}(W^*\x)=1\right)C\sqrt{2r}
\\
&\ge& \frac{\cl^{0-1}_\cd(W)-\delta}{2\|W^*\|_F} - \delta\sqrt{2r}C
\end{eqnarray*}
\proofbox

\begin{lemma}\label{lem:guarantee_given_good_init}
Fix $\epsilon >0$, $M>2$ and suppose that $\W_0$ are weights such that
\begin{itemize}
\item $\W_0\in\cw_{1.5}$
\item There is  $W^*\in M_{k,r}$ with $\|W^*\|_F\le \frac{M}{\sqrt{r}}$ such that $\cl^2_\cd(\W_0|W^*)\le \delta$ for $\delta = \frac{\epsilon}{\sqrt{2}4CM}$
\item $\cl_\cd(\W_0)\le \cl_0$
\end{itemize} 
Let $\W_1,\ldots,\W_T$ be the weights produced by algorithm \ref{alg:general_nn_training} with step size $\frac{\eta}{r}$ for $\eta \le \frac{\epsilon^2}{4M^2 \alpha^4}$, with $T\ge \frac{8M^2\cl_0}{\eta\epsilon^2}$ and with arbitrary batch size $m$. Furthermore, assume that
\[
\sqrt{r}\ge TM\eta\alpha^2 
\]
Then, there is some $t\in [T]$ such that $\E\cl^{0-1}_{\cd}(\W_t)\le 2\epsilon$.
\end{lemma}
\proof 
Since $T\le \frac{\sqrt{r}}{2\eta\alpha}$ and $T\le \frac{\sqrt{r}}{M\eta\alpha^2}$, by lemma \ref{lem:gd_change}, we have $\cl^1_{\cd}(\W_t)\le\delta$ for each $\W_t$.
Denote $\epsilon_t = \cl^{0-1}_\cd(\W_t)$ and let $\V_t$ be the stochastic gradient at time $t$. By claim \ref{claim:improvement_dir}, we have that
\begin{equation}\label{eq:grad_lower_bound}
\|\nabla \cl_{\cd}(\W_t)\|_{F}\ge \frac{(\epsilon_t-\delta)\sqrt{r}}{2M} - \delta\sqrt{2r}C \ge \frac{(\epsilon_t-\epsilon)\sqrt{r}}{2M}
\end{equation}
Now, since each $\ell_{(\x,y)}$ is $(\infty,\alpha\sqrt{r}, \alpha^2r)$-bounded (lemma \ref{lem:boundness}), we have
\[
\cl_\cd(\W_{t+1}) \le \cl_\cd(\W_{t}) - \frac{\eta}{r}\inner{\nabla \cl_{\cd}(\W_t),\V_t} + \frac{\alpha^2r}{2}\frac{\eta^2}{r^2}\|\V_t\|^2_{F}
\]
Taking expectation over the stochastic gradient and using the fact that by the boundness of $\ell_{(\x,y)}$, $\|\V_t\|_F\le \alpha\sqrt{r}$, we get
\[
\E \cl_\cd(\W_{t+1}) \le \cl_\cd(\W_{t}) - \frac{\eta}{r}\|\nabla \cl_{\cd}(\W_t)\|^2_{F} + \eta^2\frac{\alpha^4}{2}
\]
Applying equation (\ref{eq:grad_lower_bound}) we get
\[
\E \cl_\cd(\W_{t+1}) \le \cl_\cd(\W_{t}) - \frac{\eta (\epsilon_t-\epsilon)_+^2}{4M^2} + \eta^2\frac{\alpha^4}{2}
\]
Now, if $\E(\epsilon_t-\epsilon)_+^2  \ge \epsilon^2$, we have
\[
\E \cl_\cd(\W_{t+1}) \le \E \cl_\cd(\W_{t}) - \frac{\eta \epsilon^2}{4M^2} + \eta^2\frac{\alpha^4}{2}
\]
Since $\eta \le \frac{\epsilon^2}{4M^2 \alpha^4}$, we get
\[
\E \cl_\cd(\W_{t+1}) \le \E \cl_\cd(\W_{t}) - \frac{\eta \epsilon^2}{8M^2}
\]
On the other hand, we always have $\E \cl_\cd(\W_{t+1}) \ge 0$.
Hence, in the first $\frac{\cl_\cd(\W_0)8M^2 }{\eta \epsilon^2}$ steps, there must be at least one step $t$ in which $\E(\epsilon_t-\epsilon)_+^2  \le \epsilon^2$. The proof concludes as
\[
\E\epsilon_t \le \E(\epsilon_t-\epsilon)_+ + \epsilon \le \sqrt{\E(\epsilon_t-\epsilon)^2_+} + \epsilon \le 2\epsilon
\]
\proofbox

\subsection{Initial conditions}\label{sec:init}

\subsubsection{Finite support representation of kernel space functions}
For a Hilbert space $\ch$ we define by $\mathcal{B}(\ch,\reals^k)$ the collection of bounded operators from $\ch$ to $\reals^k$. Concretely, $\cb(\ch,\reals^k)$ is the collection of all functions $W:\ch\to\reals^k$ of the form $W\x=(\inner{\w_1,\x},\ldots,\inner{\w_k,\x})$ for $\w_1,\ldots,\w_k\in\ch$. $\mathcal{B}(\ch,\reals^k)$ is a Hilbert space itself w.r.t.\ the Frobenius inner product $\inner{W,W'} = \sum_{i=1}^k\inner{\w_i,\w'_i}$. For $\ba\in \reals^k$ and $\x\in\ch$ we denote by $\ba\otimes\x\in \cb(\ch,\reals^k)$ the operator $(\ba\otimes\x)(\x') = \inner{\x,\x'}\ba$. Note that $\|\ba\otimes\x\|_F = \|\ba\|\cdot\|\x\| $
Fix a normalized kernel $\kappa:\cx\times\cx\to\reals$. For $\ba\in \reals^k$ and $f:\cx\to \reals$ we define $\ba f:\cx\to\reals^k$ by $\ba f(\x) = f(\x)\ba$. For $\x\in\cx$ we denote $\kappa^\x(\x')=\kappa(\x,\x')$.
In this section we will show that functions in $\ch_\kappa^k$ can be replaced by functions of the form $\sum_{i=1}^m \ba_i \kappa^{\x_i}$ without loosing too much in terms of classification accuracy.

\begin{lemma}\label{lem:perceptron}
Let $B$ be the unit ball in a Hilbert space $\ch$. Let $\{(\x_t,y_t)\}_{t=1}^\infty \subset B\times [k] $ be a sequence of examples such that there is $W^*\in\cb(\ch,\reals^k)$ with $\|W^*\|_F\le M$ and $\forall t, \ell^{1}(W^*\x_t,y_t) = 0$.
Consider the following version of the perceptron algorithm.  Start with $W_t=0$ and for every $t$, if $(W_t\x_t)_{y_t}<a+\max_{y'\ne y_t}(W_t\x_t)_{y'}$ update $W_{t+1} = W_t+(\e_{y_t}-\e_{\hat y_t})\otimes \x_t$. Here, $\hat y_t=\argmax_{y}(W_t\x_t)_y$. Then, the algorithm makes at most $(2+2a)M^2$ mistakes. Likewise, the Frobenius norm of the final matrix is at most $(2+2a)M$
\end{lemma}
\proof
Denote $U_t = (\e_{y_t}-\e_{\hat y_t})\otimes\x_t$. We have that whenever there is an update, $\inner{W_t,U_t}\le a$. In this case
\[
\|W_{t+1}\|_F^2 = \|W_{t}+U_t\|_F^2 = \|W_t\|_F^2 + \|U_t\|^2_F + 2\inner{W_t,U_t} \le \|W_t\|_F^2 + 2 + 2a
\]
Hence, after $T$ updates, the norm of $W_t$ is at most $\sqrt{(2+2a)T}$. On the other hand $\inner{W^*,U_t} \ge 1$. Hence, after $T$ updates, the projection of $W_t$ on the direction $\frac{W^*}{\|W^*\|}$ is at least $\frac{T}{\|W^*\|}\ge \frac{T}{M}$. It follows that $\frac{T}{M}\le \sqrt{(2+2a)T}$ which implies that $T\le (2+2a)M^2$. Likewise, at this point $\|W_t\|\le \sqrt{(2+2a)T}\le (2+2a)M$.
\proofbox

\begin{corollary}\label{cor:finite_support_sep}
Suppose that $\cd$ is $M$-separable w.r.t.\ $\kappa$. Then, $\cd$ is $(4M)$-separable by a function of the form $\f^* = \sum_{i=1}^m \ba_i \kappa^{\x_i}$ with $\|\ba_i\| \le \sqrt{2}$ for all $i$ and $m\le 4M^2$
\end{corollary}
\proof (sketch)
Let $\ch$ be a Hilbert space and let $\Psi:\cx\to \ch$ be a mapping such that $\kappa(\x,\x')=\inner{\Psi(\x),\Psi(\x')}$. Let $\{(\x_t,y_t)\}_{t=1}^\infty$ be a sequence of i.i.d.\ samples from $\cd$. Suppose that we ran the algorithm from lemma \ref{lem:perceptron} with $a=1$ on the sequence $\{(\Psi(\x_t),y_t)\}_{t=1}^\infty$.
With probability 1, the number of updates the algorithm will make will be $m\le 4M^2$. Hence, upon termination, we will have $W\in \cb(\ch,\reals^k)$ such that $\cl^1_\cd(W\circ\Psi)=0$ and $W=\sum_{i=1}^m \ba_i\otimes\Psi(\x_i)$ where $\|W\|_F\le 4M$ and each $\ba_i$ is a difference of two $\e_j$'s and therefore has a norm of $\sqrt{2}$. 
The proof concludes as $W\circ\Psi = \sum_{i=1}^m \ba_i \kappa^{\x_i}$
\proofbox

\begin{lemma}\label{lem:small_norm_small_l1}
Let $B$ be the unit ball in a Hilbert space $\ch$, $\cd$ be a distribution on $B\times\cy$, $\ell:\reals^k\times\cy\to [0,\infty)$ a loss function that is convex and $L$-Lipschitz, $W^*\in \cb(\ch,\reals^k)$ and $\epsilon >0$. There are $\ba_1,\ldots,\ba_m\in \reals^k$ and $\x_1,\ldots,\x_m\in B$ such that for $W:=\sum_{i=1}^m\ba_i\otimes\x_i$ we have,
\begin{itemize}
	\item $\cl_\cd(W) \le \cl_\cd(W^*) + \epsilon$
	\item For every $i$, $ \|\ba_i\| \le \frac{\epsilon}{L}$
	\item $\|W\|_F \le \|W^*\|_F$
	\item $m\le \frac{M^2L^2}{\epsilon^2}$
\end{itemize}
\end{lemma}
\proof
Denote $M=\|W^*\|$. Suppose that we run stochastic gradient decent on $\cd$ w.r.t.\ the loss $\ell$, with
learning rate $\eta = \frac{\epsilon}{L^2}$, and with projections onto the
ball of radius $M$. Namely, we start with $W_0=0$ and at each iteration
$t\ge 1$, we sample $(\x_t,y_t)\sim\cd$ and perform the update,
\[
\tilde{W}_t = W_{t-1} - \eta\nabla\ell_{y_t}(W_{t-1}\x_t) \otimes \x_t
\]
\[
W_t = \begin{cases}
\tilde{W}_{t} & \|\tilde{W}_{t}\|_F\le M
\\
\frac{M \tilde{W}_{t}}{\|\tilde{W}_{t}\|_F} & \|\tilde{W}_{t}\|_F > M
\end{cases}
\]
Where $\nabla\ell_{y_t}(W_{t-1}\x_t)$ is a sub-gradient of  $\ell_{y_t}$ at $W_{t-1}\x_t$.
After $T=\frac{M^2L^2}{\epsilon^2}$ iterations the loss in expectation would
be at most $\epsilon$ (see for instance Chapter 14 in
\cite{shalev2014understanding}). In particular, there exists a sequence of
at most $\frac{M^2L^2}{\epsilon^2}$ gradient steps that attains a solution
$W$ with $\cl_\cd(W)\le \cl_\cd(W) + \epsilon$. 
The proof concludes as each update adds a matrix of the form
$\ba\otimes \x$ with $\|\ba\|\le L$ and possibly multiply the current matrix by a scalar of absolute value $\le 1$.
\proofbox

Similarly to corollary \ref{cor:finite_support_sep} we have
\begin{corollary}\label{cor:finite_support_non_sep}
Let $\cd$ be a distribution on $\cx\times\cy$, $\ell:\reals^k\times\cy\to [0,\infty)$ a loss function that is convex and $L$-Lipschitz, $\f^*\in \ch^k_\kappa$ and $\epsilon >0$. There are $\ba_1,\ldots,\ba_m\in \reals^k$ and $\x_1,\ldots,\x_m\in \cx$
such that for $\f:=\sum_{i=1}^m\ba_i\kappa^{\x_i}$ we have,
\begin{itemize}
	\item $\cl_\cd(\f) \le \cl_\cd(\f^*) + \epsilon$
	\item For every $i$, $ \|\ba_i\| \le \frac{\epsilon}{L}$
	\item $\|\f\|_\kappa \le \|\f^*\|_\kappa$
	\item $m\le \frac{M^2L^2}{\epsilon^2}$
\end{itemize}
\end{corollary}

\subsubsection{Initial conditions}

\begin{lemma}[\cite{vershynin2010introduction} Cor. 5.35]\label{lem:matrix_con}
Let $W \in M_{r,m}$ be a matrix with i.i.d.\ entries drawn from $\cn\left(0,\sigma^2\right)$. Then, w.p. at least $1-2\exp\left(-\frac{\alpha^2r}{2}\right)$, $\|W\|_{2}\le (1+\alpha)\sigma(\sqrt{r}+\sqrt{m})$
\end{lemma}

\begin{corollary}\label{cor:bounded_norm}
Suppose that $r\gtrsim \log\left(\frac{|S|}{\delta}\right) + d$. Then, 
\begin{itemize}
\item W.p.\ $\ge 1-\delta$ we have that $\W_0\in \cw'_{1.5}$.
\item If also $r\ge k$ then w.p.\ $\ge 1-\delta$ we have that $\W_0\in \cw_{1.5}$.
\end{itemize}
\end{corollary}

\begin{theorem}[\cite{daniely2016toward}]\label{thm:dfs_ker}
Let $\cs$ be a skeleton with $C$-bounded activations.
Let $\W$ be a random initialization 
$$r \ge \frac
	{(4C^4)^{\depth(\cs)+1} \log\left({8|\cs|}/{\delta}\right)}
	{\epsilon^2} \,.$$
Then, for all $\x,\x'$, with probability of at least $1-\delta$,
$$|k_{\W}(\x,\x')-k_{\cs}(\x,\x')|\le \epsilon\,.$$
\end{theorem}

\begin{theorem}[\cite{daniely2016toward}]\label{thm:dfs_ker_relu}
Let $\cs$ be a skeleton with ReLU activations.
Let $\W$ be a random initialization with
$$r \gtrsim \frac
	{\depth^2(\cs)\log\left({|\cs|}/{\delta}\right)}
	{\epsilon^2} \,.$$
Then, for all $\x,\x'$ and $\epsilon \lesssim \frac{1}{\depth(\cs)}$, with probability of at least $1-\delta$,
$$|k_{\W}(\x,\x')-k_{\cs}(\x,\x')|\le \epsilon\,.$$
\end{theorem}

\begin{theorem}[\cite{daniely2016toward}]\label{thm:dfs_dist}
Let $\cs$ be a skeleton with $C$-bounded activations. Let $\bh^*\in \ch^k_{\cs}$ with $\|\bh^*\|_\cs\le M$. Suppose
$$r \gtrsim \frac
	{L^4 \, M^4 \, (4C^4)^{\depth(\cs)+1}
		\log\left(\frac{LMC |\cs|}{\epsilon\delta}\right)}
	{\epsilon^4} \,.$$
Then, with probability of at least $1-\delta$ over the choices
of $\W_0$ there is a prediction matrix $W^*$ such that $\|W^*\|_F\le \frac{2M}{\sqrt{r}}$ and $\cl_\cd(\W_0|W^*)\le \cl_\cd(\bh^*) + \epsilon$
\end{theorem}
\begin{theorem}[\cite{daniely2016toward}]\label{thm:dfs_dist_relu}
Let $\cs$ be a skeleton with ReLU activations and
$\epsilon\lesssim {1}/{\depth(\cs)}$. Let $\bh^*\in \ch^k_{\cs}$ with $\|\bh^*\|_\cs\le M$. Suppose
$$r \gtrsim \frac{L^4 \,M^4\, \depth^3(\cs)\,
	\log\left(\frac{LM |\cs|}{\epsilon\delta}\right)}{\epsilon^4}+d \,.
$$ 
Then, with probability of at least $1-\delta$ over the choices
of $\W_0$ there is a prediction matrix $W^*$ such that $\|W^*\|_F\le \frac{2M}{\sqrt{r}}$ and $\cl_\cd(\W_0|W^*)\le \cl_\cd(\bh^*) + \epsilon$
\end{theorem}
Theorems \ref{thm:dfs_dist} and \ref{thm:dfs_dist_relu} are similar, but not identical, to theorems 4 and 5 from \cite{daniely2016toward}. We next prove theorem \ref{thm:dfs_dist_relu}. The proof of theorem \ref{thm:dfs_dist} is very similar.
\proof (sketch)
By corollary \ref{cor:finite_support_non_sep} we can restrict to the case that $\bh^* = \sum_{i=1}^m \ba_i\kappa_{\cs}^{\x_i}$ such that $\forall i, \|\ba_i\|\le \frac{\epsilon}{L}$ and $m\le \frac{M^2L^2}{\epsilon^2}$. Define $W^* = \sum_{i=1}^M \ba_i\otimes \frac{R_{\x_i}(\W_0)}{r}$ and
\[
\bh(\x) = W^*R_\x(\W_0) = \sum_{i=1}^m \ba_i\kappa_{\W_0}^{\x_i}(\x)
\]
Denote $\alpha = (\sqrt{18})^{\depth{\cs}}\sqrt{\comp(\cs)}$. We will show that the conclusion of the theorem holds given the following three conditions, that by theorem \ref{thm:dfs_ker_relu} and corollary \ref{cor:bounded_norm} happens w.p. $\ge 1-\delta$ over the choice of $\W_0$.
\begin{itemize}
\item For $\left(1-\frac{\epsilon}{(1+\sqrt{2}\alpha)ML}\right)$-fraction (according to $\cd$) of the examples, 
\begin{equation}\label{eq:most_exam}
\forall i, |\kappa_{\W_0}(\x_i,\x)-\kappa_{\cs}(\x_i,\x)|\le \frac{\epsilon^2}{M^2L^2}
\end{equation}
\item For all $i,j$, $|\kappa_{\W_0}(\x_i,\x_j)-\kappa_{\cs}(\x_i,\x_j)|\le \frac{\epsilon^2}{M^2L^2}$ 
\item $\W_0\in\cw'_{1.5}$
\end{itemize}
First, we have
\[
\|W^*\|^2_F = \frac{1}{r}\sum_{i,j=1}^m \inner{\ba_i,\ba_j}\kappa_{\W_0}(\x_i,\x_j)
\]
Since $|\kappa_{\W_0}(\x_i,\x_j)-\kappa_{\cs}(\x_i,\x_j)|\le \frac{\epsilon^2}{M^2L^2}$ for all $i,j$ and also $|\inner{\ba_i,\ba_j}|\le\frac{\epsilon^2}{L^2}$ we have
\begin{eqnarray*}
r\|W^*\|^2_F &\le&  \left[\sum_{i,j=1}^m \inner{\ba_i,\ba_j}\kappa_{\cs}(\x_i,\x_j)\right]  + m^2\frac{\epsilon^4}{M^2L^4}
\\
&=& \|\bh^*\|_\cs^2 + m^2\frac{\epsilon^4}{M^2L^4} \le 2M^2
\end{eqnarray*}
Now, for the examples satisfying $\forall i, |\kappa_{\W_0}(\x_i,\x)-\kappa_{\cs}(\x_i,\x)|\le \frac{\epsilon^2}{M^2L^2}$ we have
\[
\|\bh(\x)-\bh^*(\x)\| 
= \left\|\sum_{i=1}^m \ba_i(\kappa_{\W_0}(\x_i,\x)-\kappa_{\cs}(\x_i,\x)) \right\| 
\le \sum_{i=1}^m\|\ba_i\|\frac{\epsilon^2}{M^2L^2} \le m\frac{\epsilon}{L}\frac{\epsilon^2}{M^2L^2} \le \frac{\epsilon}{L}
\]
Since the loss is $L$-Lipschitz, it grows by at most $\epsilon$ on these examples, when we move from $\bh^*$ to $\bh$. As for the remaining examples, since $\W_0\in\cw'_{1.5}$ and $\|W^*\|_F\le \frac{\sqrt{2}M}{\sqrt{r}}$, lemma \ref{lem:boundness_relu} implies that 
\[
\|\bh(\x)-\bh^*(\x)\| \le \|\bh(\x)\| + \|\bh^*(\x)\| \le \sqrt{2}\alpha M + M
\]
Hence, when we move from $\bh^*$ to $\bh$, the loss grows by at most $(1+\sqrt{2}\alpha)ML$. As the remaining examples occupies at most $\frac{\epsilon}{(1+\sqrt{2}\alpha)ML}$-fraction of the examples, these examples contributes at most $\epsilon$ to the loss when moving from $\bh^*$ to $\bh$.
\proofbox

\begin{corollary}\label{cor:init_sep}
Let $\cs$ be a skeleton with $C$-bounded activations.
Suppose that $\cd$ is $M$-separable w.r.t.\ $\kappa_\cs$, $r\gtrsim (4C^4)^{\depth(\cs)+1}M^4\log\left(\frac{|\cs|M}{\delta}\right)$. Then, w.p. $\ge 1-\delta$ over the choice of $\W_0$, there is $W^*\in M_{k,r}$ with $\|W^*\|_F\le \frac{10M}{\sqrt{r}}$ such that $\cl^1_{\cd}(\W_0|W^*)\le \delta$
\end{corollary}
\proof  (sketch)
By corollary \ref{cor:finite_support_sep}, $\cd$ is separable by a function $\bh^* = \sum_{i=1}^m \ba_i\kappa_{\cs}^{\x_i}$ such that $\forall i, \|\ba_i\|\le \sqrt{2}$, $m\le 4M^2$ and $\|\bh^*\|\le 4M$.
Define $W = \sum_{i=1}^M \ba_i\otimes \frac{R_{\x_i}(\W_0)}{r}$ and
\[
\bh(\x) = WR_\x(\W_0) = \sum_{i=1}^m \ba_i\kappa_{\W_0}^{\x_i}(\x)
\]
Now, for $r\gtrsim (4C^4)^{\depth(\cs)+1}M^4\log\left(\frac{|\cs|M}{\delta}\right)$ we have by theorem \ref{thm:dfs_ker} that w.p. $\ge 1-\frac{\delta}{2}$, $|\kappa_{\W_0}(\x_i,\x_j)-\kappa_{\cs}(\x_i,\x_j)|\le \frac{1}{32M^2}$ for all $i,j$ and also $|\inner{\ba_i,\ba_j}|\le 2$. Hence,
\begin{eqnarray*}
r\|W\|^2_F &\le&  \left[\sum_{i,j=1}^m \inner{\ba_i,\ba_j}\kappa_{\cs}(\x_i,\x_j)\right]  + 2m^2\frac{1}{32M^2}
\\
&=& \|\bh^*\|_\cs^2 + 2m^2\frac{1}{32M^2} \le \|\bh^*\|_\cs^2 + 32M^4\frac{1}{32M^2} \le 17M^2 \le 25 M^2
\end{eqnarray*}
Now, w.p.\ $\ge 1-\frac{\delta}{2}$, $(1-\delta)$-fraction of the examples satisfies $|\kappa_{\W_0}(\x_i,\x)-\kappa_{\cs}(\x_i,\x)|\le \frac{1}{16\sqrt{2}M^2}$ for all $i$. For those examples we have
\[
\|\bh(\x)-\bh^*(\x)\| 
= \left\|\sum_{i=1}^m \ba_i(\kappa_{\W_0}(\x_i,\x)-\kappa_{\cs}(\x_i,\x)) \right\| 
\le \sum_{i=1}^m\sqrt{2}\frac{1}{16\sqrt{2}M^2} \le \frac{4M^2\sqrt{2}}{16\sqrt{2}M^2} = \frac{1}{4}
\]
Those examples satisfies $\ell^{\frac{1}{2}}(\bh(\x),y)=0$, and hence $\cl^{\frac{1}{2}}_\cd(h)\le \delta$. The proof concludes by taking $W^*=2W$
\proofbox

\begin{lemma}[e.g.\ \cite{kamathbounds}]\label{lem:Gauss_bound}
Let $X_1,\ldots,X_k$ be independent standard Gaussians. Then $\E\max_iX_i\le \sqrt{2\log(k)}$
\end{lemma}
\proof
Denote $Z = \max_iX_i$ and let $t=\sqrt{2\log(k)}$. By Jensen's inequality and the moment generating function of the normal distribution we have that
\[
e^{t\E Z} \le \E e^{tZ} = \E \max_i e^{tX_i} \le k\E e^{tX_1} = ke^{\frac{t^2}{2}} 
\]
Hence, $\E Z\le \frac{\log(k)}{t} + \frac{t}{2} = \sqrt{2\log(k)}$
\proofbox

\begin{lemma}\label{lem:bound_init_loss}
If all activations are $C$ bounded then $\E_{\cw_0}\cl_\cd(\W_0) \le (1+C\sqrt{2})\log(k)$
\end{lemma}
\proof
Fix $\z=(\x,y)$. We have that the distribution of $\ell_\z(\W_0)= \ell(W^\pred_0R_\x(\W_0),y)$ is the distribution of
$\ell((X_1,\ldots,X_k),1)$ where $X_1,\ldots,X_k$ are independent Gaussians of mean zero and variance $\sigma^2 = \frac{\|R_\x(\W_0)\|^2}{r} \le C^2$. Hence we have that
\begin{eqnarray*}
\E_{\W_0}l_\z(\W_0) &=& \E_{X}-\log\left(\frac{e^{X_1}}{\sum_{i=1}^ke^{X_i}}\right)
\\
&=& -\E X_1 + \E\log\left(\sum_{i=1}^ke^{X_i}\right)
\\
&=& \E\log\left(\sum_{i=1}^ke^{X_i}\right)
\\
&\le& \E\log\left(k e^{\max_iX_i}\right)
\\
&\le& \log(k)+\E\max_iX_i
\end{eqnarray*}
and the proof concludes as by lemma \ref{lem:Gauss_bound} $\E\max_iX_i\le \sigma\sqrt{2\log(k)}$
\proofbox

\paragraph{Derivation of the main theorems}. Theorem \ref{thm:main_bounded_no_sep} follows from theorem \ref{thm:dfs_dist} and lemma \ref{lem:guarantee_given_good_init_relu}. Likewise, theorem \ref{thm:main_relu_no_sep} follows from theorem \ref{thm:dfs_dist_relu} and lemma \ref{lem:guarantee_given_good_init_relu}. Finally, theorem \ref{thm:main_bounded_sep} follows from corollary \ref{cor:init_sep}, lemma \ref{lem:bound_init_loss} and lemma \ref{lem:guarantee_given_good_init}.

\paragraph{Acknowledgements:}
The author thanks Roy Frostig, Yoram Singer and Kunal Talwar for valuable discussions and comments.

\bibliographystyle{plainnat}

\bibliography{bib}

\end{document}